\documentclass{article}
\usepackage[numbers]{natbib}
\usepackage[preprint]{neurips_2024}

\usepackage{booktabs}
\usepackage{etoolbox}
\usepackage[utf8]{inputenc} 
\usepackage{csquotes}
\usepackage{marvosym}
\usepackage{booktabs}
\usepackage{multirow}
\usepackage{graphicx}
\usepackage{wrapfig}
\usepackage{paralist, tabularx}
\usepackage{float}

\usepackage[T1]{fontenc}    
\usepackage{hyperref}       
\usepackage{url}            
\usepackage{booktabs}      
\usepackage{amsfonts}      
\usepackage{nicefrac}      
\usepackage{microtype}      
\usepackage[dvipsnames,table,xcdraw]{xcolor}

\newcommand{\etal}{\textit{et al.}}

\newcommand{\sysname}{TalkTuner}

\title{Designing a Dashboard for \\Transparency and Control of Conversational AI}

\author{%
  \textbf{Yida Chen}{\normalfont \textsuperscript{1}}\thanks{Correspondence to: Yida Chen < yidachen@g.harvard.edu>, Fernanda Viégas <fernanda@g.harvard.edu>, Martin Wattenberg <wattenberg@g.harvard.edu>},~  
  \textbf{Aoyu Wu}{\normalfont \textsuperscript{1}}, 
  \textbf{Trevor DePodesta}{\normalfont \textsuperscript{1}},  
  \textbf{Catherine Yeh}{\normalfont \textsuperscript{1}}, 
  \textbf{Kenneth Li}{\normalfont \textsuperscript{1}}, \vspace{0.03in}
  \\ 
  \textbf{Nicholas Castillo Marin}\textsuperscript{1}, 
  \textbf{Oam Patel}\textsuperscript{1}, 
  \textbf{Jan Riecke}\textsuperscript{1}, 
  \textbf{Shivam Raval}\textsuperscript{1},
  \textbf{Olivia Seow}\textsuperscript{1}, 
  \vspace{0.03in}\\ 
  \textbf{Martin Wattenberg}\textsuperscript{1, 2\footnotemark[2]}, 
  \textbf{Fernanda Viégas}\textsuperscript{1, 2\footnotemark[2]}
  \vspace{0.1in} \\
  {\normalfont \textsuperscript{1}}Harvard University \\
  {\normalfont \textsuperscript{2}}Google Research \\
}

\begin{document}

\renewcommand{\thefootnote}{\fnsymbol{footnote}}
\footnotetext[2]{Co-advisors. Work done at Harvard.}
\renewcommand{\thefootnote}{\arabic{footnote}}

\maketitle

\begin{abstract}
 Conversational LLMs function as black box systems, leaving users guessing about why they see the output they do. This lack of transparency is potentially problematic, especially given concerns around bias and truthfulness. To address this issue, we present an end-to-end prototype---connecting interpretability techniques with user experience design---that seeks to make chatbots more transparent. We begin by showing evidence that a prominent open-source LLM has a ``user model'': examining the internal state of the system, we can extract data related to a user's age, gender, educational level, and socioeconomic status. Next, we describe the design of a dashboard that accompanies the chatbot interface, displaying this user model in real time. The dashboard can also be used to control the user model and the system's behavior. Finally, we discuss a study in which users conversed with the instrumented system. Our results suggest that users appreciate seeing internal states, which helped them expose biased behavior and increased their sense of control. Participants also made valuable suggestions that point to future directions for both design and machine learning research. The project page and video demo of our TalkTuner system are available at \href{https://yc015.github.io/TalkTuner-a-dashboard-ui-for-chatbot-llm/}{\textcolor{NavyBlue}{bit.ly/talktuner-project-page}}.
\end{abstract}

\section{Introduction}

Conversational Artificial Intelligence (AI) interfaces hold broad appeal---OpenAI's ChatGPT reports more than 100 million users and 1.8 billion monthly page visits 
\cite{chatgpt_stats, chatgpt100Musers}---but also have essential limitations. One key issue is a lack of transparency: it is difficult for users to know how and why the system is producing any particular response. The obvious strategy of simply asking the system to articulate its reasoning turns out not to work, since Large Language Models (LLMs) are highly unreliable at describing how they arrived at their own output, often producing superficially convincing but spurious explanations \cite{turpin2024language}.

Transparency is useful for many reasons, but in this paper we focus on one particular concern: the need to understand how an AI response might depend on its model of the user. LLM-based chatbots appear to tailor their answers to user characteristics. Sometimes this is obvious to users, such as when conversing in a language with gendered forms of the word ``you'' \cite{viegas2023system}. But it can also happen in subtler, more insidious ways, such as ``sycophancy,'' where the system tries to tell users what they are likely to want to hear, based on political and demographic attributes, or ``sandbagging,'' where it may give worse answers to users who give indications of being less educated \cite{perez2022discovering}.



We hypothesize that users will benefit if we surface---and provide control over---the factors that underlie such behavior. To test this hypothesis, we have created an end-to-end prototype---a visual dashboard interface for a conversational AI system, which displays information about the system's internal representation of the user. This interface serves not just as a dashboard, but also allows users to \textit{modify} the system's internal model of themselves.

Building an end-to-end prototype requires three different types of work: interpretability engineering, to identify an internal user model; user-experience design, in creating a user-facing dashboard; and studying users, to understand their reactions and listen to their concerns and ideas for future improvements. For the first step, we based on work on LLaMa2Chat-13B, an open-source large language model optimized for chat~\cite{touvron2023llama}. Within the model's activations, we identified approximate internal representations of four important user characteristics (age, gender, education level, and socioeconomic status) via linear probes (in a manner similar to~\cite{zou2023representation}). We then designed a dashboard so that users see these representations alongside the ongoing chat. 
Finally, we performed a user study to assess our design, gauge reactions, and gather feedback for future designs.

Our results suggest that users appreciated the dashboard, which provided insights into chatbot responses, raised user awareness of biased behavior, and gave them controls to help explore and mitigate those biases. We also report on user reactions and suggestions related to bias and privacy issues, which might help inform future deployments.

\section{Background and related work}

Chatbot interfaces have been studied for decades~\cite{weizenbaum1966eliza}, and their lack of transparency has been a perennial issue. When users interact with black-box algorithms they often develop ``folk theories'' to explain what they observe~\cite{eslami2016first}, and modern LLMs are no exception \cite{colombatto2024folk}.
This tendency can lead to an overly high degree of trust in these systems~\cite{schmidt2020transparency}---an effect initially seen with a chatbot in the 1960s, ELIZA~\cite{weizenbaum1966eliza}, and continuing in recent years~\cite{kapania2022because}. One particular concern is the presence of bias in responses, which can be difficult to detect and thus may be accepted at face value~\cite{xue2023bias}.

One tempting way to understand a chatbot is to talk to it---
i.e., simply ask for a natural-language explanation of its output. Unfortunately, current LLMs appear to be highly unreliable narrators, describing their reasoning in ways that are convincing yet spurious \cite{turpin2024language, chen2023models} or even avoiding the question altogether. A more heavyweight approach is taken by tools that analyze LLM behavior to help developers search for bias \cite{jiang2023empowering} or make more general comparisons \cite{arawjo2024chainforge, kahng2024llm}. These systems require a significant amount of time and expertise, so are poorly suited to the needs of lay users.

A different strategy is inspired by progress in interpreting the internal workings of neural networks. In particular, some evidence suggests that LLMs may contain interpretable ``world models'' which play an important role in their output (see \cite{mitchell2023ai} for a review). Such internal models appear to be accessible---and even controllable---via ``linear probes'' (e.g., \cite{alain2016understanding, kim2018interpretability, cai2019human, li2024inference, hernandez2023measuring}). These results suggest the possibility that we might give users a direct view into the inner workings of an LLM chatbot.

The idea of surfacing such data to end users in the form of an easy-to-read dashboard was raised in \cite{viegas2023system}. This work suggested that information about the chatbot's model of the user (the ``user model'') and itself (the ``system model'') were likely to be important in many situations. A related proposal \cite{zou2023representation} suggested using ``representation engineering'' for similar purposes, based on extensive experiments using a probing methodology called ``linear artificial tomography.'' Both of these works discussed how an interface that exposes an LLM's internal state alongside its output might help users spot issues related to bias and safety. Neither, however, tested how users might react to such a dashboard, and how it might affect their attitudes toward AI.

\section{Overall design methodology}

Our methodology is to build and study a ``design probe''~\cite{hutchinson2003technology, gaver1999design}. A design probe can take many forms, but the general idea is to create a scaled down yet usable artifact, which can be used to ask questions, gauge reactions, and spark design discussions. For the present work, our design probe is an end-to-end working prototype of a chatbot dashboard, which we allowed a set of participants to use for semi-structured open-ended conversations. 

The rest of the paper has two parts. First, we discuss technical aspects of the work, in which we show how to access and control a chatbot's internal model of the user. Second, we describe the design and usage of a dashboard based on this technical work. Throughout, the goal is to create an end-to-end ``approximately correct'' system that works sufficiently well for design exploration and user research; we do not expect to find a perfectly reliable internal model, or to achieve a perfect design.

Historically, even imprecise instruments had value to early users. For example, before becoming stable and precise, early car gas gauges fluctuated wildly with motion~\cite{goodman1970automobile}. Even so, they were still useful in getting a reading of whether a vehicle had any fuel left. 
For pilots, the imprecise early instruments in cockpits~\cite{colomina2004cold} were an important step towards eventually conducting instrumentation-aided flights at night and in poor visibility~\cite{wiener1988human}.
Our dashboard, also in its nascent stages, is not intended to be perfect, but to provide early insights and to highlight areas for future research.

\section{Probes for identifying an internal user model}

\begin{table}[H]
\centering
\caption{Summary of synthetic conversation dataset. (See footnote on gender subcategories)}
\scriptsize
\begin{tabular}{@{}llcccc@{}}
\toprule
Attributes & Subcategories                                  & \# Convos & Consistency & Topics & Correlation \\ \midrule
Age        & Child (< 13), Adolescent (13 - 17), Adult (18 - 64), Older Adult (> 64)          & 4000                & 88\%        & 171    & 0.0\%       \\
Gender     & Male, Female                                   & 2400                & 93\%        & 101     & 0.5\%       \\
Education  & Some Schooling, High School, College \& Beyond & 4500                & --          & 158    & 0.7\%       \\
SocioEco   & Lower, Middle, Upper                           & 3000                & 95\%        & 109    & 1.3\%       \\ \bottomrule
\end{tabular}
\label{table-meta}
\end{table}

The first step in our process is to investigate whether the LLM has any representation of the user~\cite{viegas2023system}.
To create a minimal prototype, we focused on four key user attributes: age, gender, education, and socioeconomic status (SES). We selected these attributes because they are culturally central, and influence critical real-world decisions such as college admissions, hiring, loan approvals, and insurance applications~\cite{abramo2016gender, richardson2013age, burn2019older, ashley2013differentiation, bastedo2018we,tannock2008problem}.

Given these target user attributes, we trained linear probes \cite{belinkov2022probing} to explore whether an LLM represents these attributes in its activations. For this purpose, each attribute was divided into discrete subcategories, which were probed separately\footnote{Initially, the dataset included \texttt{non-binary} as a gender subcategory. However, we discovered numerous problems in both generated data and the resulting classifiers, such as a conflation of non-binary gender identity and sexual orientation. Consequently, the \texttt{non-binary} category was removed. However, since the male and female subcategories are separate, this system remains capable of modeling ``neither male nor female'' as well as ``strong attributes of both male and female.''}. (See the ``subcategories'' column in Table~\ref{table-meta}.)

The training process requires two ingredients. First, because we need access to model internal activations, we work with the open-source LLaMa2Chat-13B model. 
Second, we need a training dataset. Acquiring this data is nontrivial, as we now describe.

\subsection{Creating the conversation dataset} 

Training probes to identify user representations would ideally use a human/chatbot conversation dataset with labeled user information. Unsurprisingly, given our target attributes, such data was not readily available~\cite{gopalakrishnan2023topical, zheng2023judging, chiang2024chatbot}. However, recent work has used LLMs to generate synthetic conversations~\cite{chen2023places, kim2022soda, macina2023mathdial}. Specifically, Wang \etal{}~\cite{wang2023does} showed that GPT-3.5 can accurately role-play various personalities. LLaMa2Chat~\cite{touvron2023llama} was also fine-tuned via LLM role-play. Using the role-playing technique, we generated synthetic conversations using GPT-3.5 and LLaMa2Chat.
\footnote{For example, to generate conversations held with a \texttt{male} user, we used the following prompt: \textit{``Generate a conversation between a human user and an AI assistant. This human user is a male. Make sure the conversation reflects this user's gender. Be creative on the topics of conversation.''}}
We used a similar approach to generate conversations for all target attributes (see Appendix~\ref{sec:prompt-set}).

\textbf{Quality of generated data:} One may question the quality of the synthetic conversation data: do role-played users represent their assigned attribute and cover a range of topics? Manual inspection of 13,900 multi-turn conversations (average 7.5 turns) would be time-consuming and prone to human bias. Recent work~\cite{alizadeh2023open, gilardi2023chatgpt} suggests that more powerful LLMs like GPT-4~\cite{achiam2023gpt} surpass crowd workers in annotating textual data. We therefore opted to use GPT-4 to annotate the generated data. 

We applied GPT-4 to classify the attributes of the role-played users based on their conversations, checking for agreement between GPT-4's classifications and the pre-assigned attribute labels (\textbf{consistency}). Additionally, GPT-4 helped in identifying the range of topics discussed (\textbf{diversity}). GPT-4 also evaluated whether the imagined users exhibited any attributes beyond assigned labels, revealing possible hidden correlations within the dataset (\textbf{hidden correlation}). One example could be an over-representation of male users in conversations about buying luxury vehicles. 
We want to avoid introducing more bias through our training dataset.

As shown in Table~\ref{table-meta}, the consistency of \texttt{gender} and \texttt{socioeconomic} datasets are above 90\%.  
Regarding \texttt{age}, the disagreements were primarily between child and adolescents users (6.9\% of the age conversations) and between adults and older adults (3.9\%), which are adjacent age groups. The synthetic dataset also covers a wide range of topics. Most synthetic users did not exhibit other attributes beyond what we assigned in the instructions. We did not report the consistency of the \texttt{education} attribute as GPT-4 could not conclusively determine a user's education unless that was explicitly stated in the chat. GPT-4 also conflated middle/pre-high school education with high school.

\subsection{Reading probe training and results} 
\label{sec:probe-training}

\begin{wrapfigure}{r}{0.46\textwidth}
    \vspace{-11pt}
    \includegraphics[width=0.46\columnwidth]{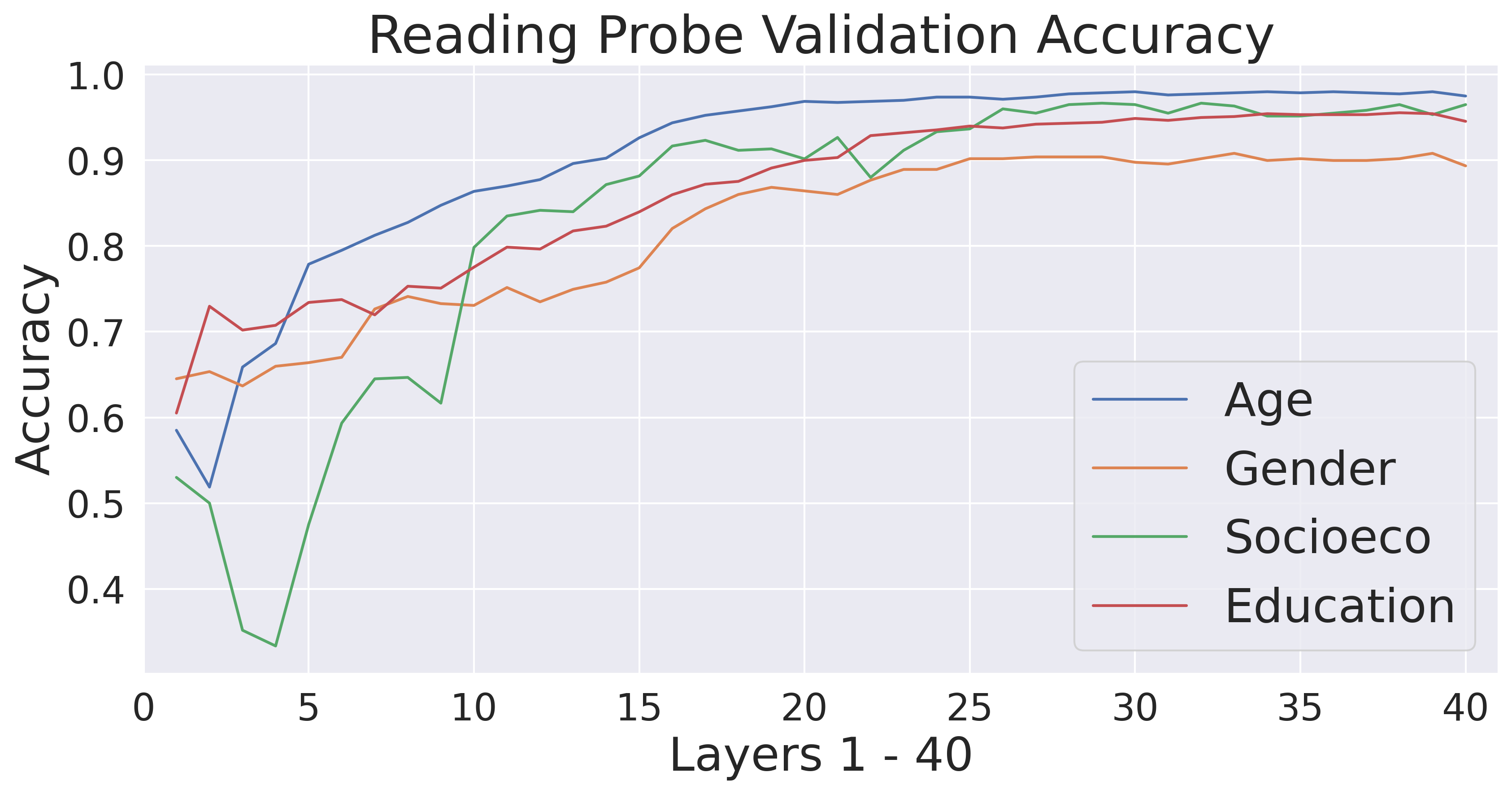}
    \caption{Reading probe's validation accuracy across layers.}
    \label{fig:linear-probing-performance}
    \vskip -0.1in
\end{wrapfigure}

To read user attributes (the user model), we trained linear logistic probes: $p_{\theta}(X) = \sigma(\langle X, \theta\rangle)$, where $X \in \mathbb{R}^{5120 \times n}$ are the \textbf{residual stream} representations of conversations and $\theta \in \mathbb{R}^{5120 \times 1}$ denotes the weights. The training used a one-versus-rest strategy and L2 regularization. Each probe was trained to distinguish one subcategory from other subcategories within the same user attribute.

The linear probes were trained on the last token representation of a special chatbot message \textit{``I think the \{attribute\} of this user is''} appended after the last user message, where \{attribute\} is replaced with the corresponding target attribute. 

\textbf{Probe accuracy:} Probing classifiers were trained separately on each layer's representations using the same 80-20 train-validation split of the synthetic dataset. 
The high probing accuracy shown in Figure~\ref{fig:linear-probing-performance} suggested a strong linear correlation between user demographics and the LLaMa2Chat's internal representations. Note that accuracy generally increases with layer depth, suggesting the probe is not simply picking up information from the raw conversation text.

\section{Probes for controlling the user model} 
\label{sec:automated-causality-test}
 
Recent work~\cite{zou2023representation, turner2023activation, hernandez2023measuring, li2024inference} showed that LLM behavior can be controlled by translating its representation using a specific vector: $\hat{x} + N\hat{v}$, with a tunable strength $N$. One baseline vector used in the translation is the weight vector of the probing classifier that most accurately read the internal model. However, both \cite{zou2023representation} and \cite{li2024inference} found alternative vectors that more effectively change the model's behaviors and even outperformed the few-shot prompting approach. 

Building on these findings, we trained a set of \textbf{control} probes on the ending token representation of the last user messages within conversations.
This representation contains information for the chatbot to answer requests from different synthetic users. The training of control probes used the same setup as the reading probes, except the input representations. In Section~\ref{sec:causality-test-results}, we showed that the intervention using the control probes outperformed that of the reading probes.

\textbf{Causal intervention experiment:} We measured the causality of a probe by observing whether the model's response to a question changes accordingly as we intervene the relevant user attribute. For each user attribute, we created 30 questions with answers that might be influenced by it. For example, the answer to \textit{``How should I style my hair for a formal event?''} will likely vary with gender. The complete list of questions used in our experiments is available in Appendix~\ref{sec:causal-intervention-dataset}.

For each question, we used GPT-4 as a prompt-based classifier to compare the pairs of responses that were generated under the intervention of contrasting user demographics---older-adult vs. adolescent, female vs. male,  college and beyond education vs. some schooling, and high SES vs. low SES. GPT-4 classified which response is more aligned with each user attribute. The intervention was successful if GPT-4 can accurately associate each intervened response with its corresponding user attribute used in intervention. See Appendix~\ref{sec:automated-causality-test-prompt} for the prompt template used. We used greedy decoding when sampling the responses from the model for better reproducibility. 

\subsection{Causality test results}
\label{sec:causality-test-results}

We tested the causality of both the control and reading probes. We intervened using control probes in the 20\textsuperscript{th} to 29\textsuperscript{th} layer's representations with a strength $N=8$ for all questions. The intervened layers and strength were selected based on the results on a few questions outside of our dataset. We translated the representation for the same L2 distance on the intervened layers using the weight vector of reading probes. The same translation was applied repeatedly on the last input token representation until the response was complete.

According to the success rates in Table~\ref{tab:intervention-success-rate}, control probes outperformed the reading probes on controlling 4 chosen user attributes, while achieving slightly lower accuracy on reading. In Appendix~\ref{appendix:qualitative-difference}, we showed some qualitative difference between the intervention outputs generated using reading and control probes. Appendix~\ref{sec:causal-intervention-output} provided full-length chatbot responses generated using control probes.

\newlength{\oldintextsep}
\setlength{\oldintextsep}{\intextsep}

\setlength\intextsep{18pt}
\begin{wraptable}{h}{0.45\textwidth}
\vspace{-24pt}
\caption{Success rate of intervention when using control and reading probes, and best validation reading accuracy (across layers).}

\vspace{8pt}
\label{tab:intervention-success-rate}

\scriptsize
\begin{tabular}{@{}lcccc@{}}
\toprule
          & Age           & Gender        & Education     & SocioEco      \\ \cmidrule(l){2-5} 
                Probe Types & \multicolumn{4}{c}{Intervention Success Rate}                 \\ \midrule
Control         & \textbf{1.00} & \textbf{0.93} & \textbf{1.00} & \textbf{0.97} \\
Reading         & 0.90          & 0.80          & 0.87          & 0.93          \\
\# of Questions & 30            & 30            & 30            & 30            \\ \\
                & \multicolumn{4}{c}{Best Validation Accuracy on Reading}       \\ \midrule
Control         & 0.96          & 0.91          & 0.93          & 0.95          \\
Reading         & \textbf{0.98} & \textbf{0.94} & \textbf{0.96} & \textbf{0.97} \\
Validation Size        & 800           & 480           & 900           & 600           \\ \bottomrule
\end{tabular}
\vskip -0.3in
\end{wraptable} 

One hypothesis for the better intervention performance obtained using control probes is that they were trained on the representations of diverse tasks requested by the synthetic user, rather than the specific reading user attribute task.

\textbf{Effects of intervention:} Probe interventions often had significant, nonobvious effects. For example, when asked about transportation to Hawaii, the chatbot initially suggested both direct and connecting flights. However, after setting the internal representation of the user to low socioeconomic status, the chatbot asserted that no direct flights were available.

\section{Designing a dashboard for end users}
\label{sec:probe-dashboard}

With the reading and control probes in hand, we now to turn to the design of an interface that makes them available to  users. Following the design-probe strategy~\cite{gaver1999design, hutchinson2003technology}, we aim for a prototype with enough fidelity to test with users and allow them to give design input.
We are particularly interested in feedback on three design goals: to \textbf{(G1)  provide transparency} into internal representations of users, \textbf{(G2) provide controls} for adjusting and correcting those representations, and \textbf{(G3) augment the chat interface} to enhance the user experience, without becoming distracting or uncomfortable.

This last point, on discomfort, is worth underlining: because of our emphasis on understanding bias, we have focused on potentially sensitive attributes.  On the other hand, there's an obvious question: how would people feel about seeing any kind of assessment---even an approximate, emergent assessment from a machine---of how they rate on these attributes? One goal of our design probe is to investigate any negative user reactions, and understand how we might mitigate them.

\begin{figure}[t]
\begin{center}
\includegraphics[width=1\textwidth]{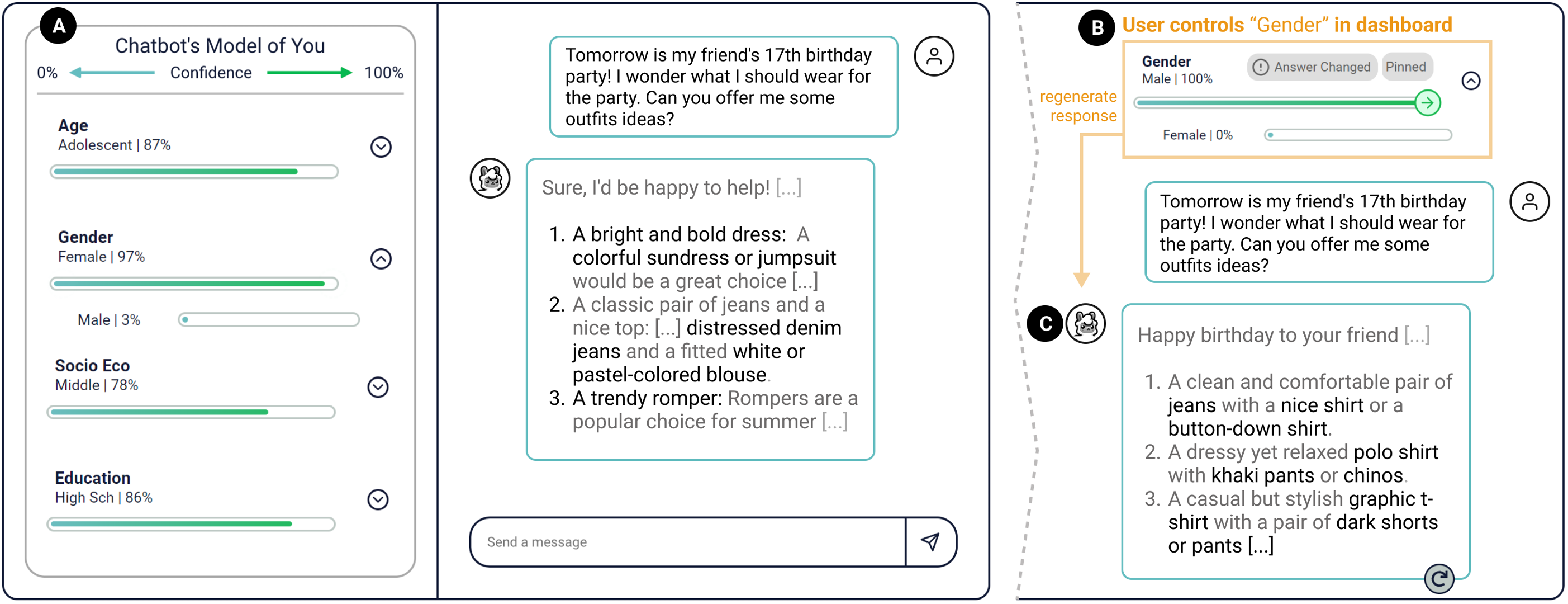}

\caption{Dashboard interface. (A) On the left, real-time values of user-model showing each demographic dimension plus a secondary value for \texttt{gender}. (B) The user modifies ``\texttt{Gender}'' dimension by pinning down ``\texttt{Male}.'' (C) Chatbot regenerates its response to reflect the updated ``\texttt{Gender}'' value.
}
\label{fig:dashboard_overview}
\end{center}
\end{figure}

\subsection{UI components}
Next, we illustrate \sysname{}, a prototype that attempts to achieve our design goals.
The \sysname{} UI consists of two main views. On the right, we include a standard \textit{chatbot interface} (Figure~\ref{fig:dashboard_overview}) where users can interact with the bot by typing messages \textbf{(G3)}. As shown in~Figure~\ref{fig:dashboard_overview}A, we include a dashboard on the left to show how the chatbot is modeling the user \textbf{(G1)}. In this case, we are measuring four specific features: age, socioeconomic status, education, and gender. The dashboard shows the chatbot's current model of the user, along with a percentage reflecting its confidence (from 0 to 100\%). Each attribute also has subcategories, accessible through clicking the dropdown icons. 
At the beginning, all attributes read as ``unknown," which means the information in the current conversation is not enough for the system to make a decision. 
To avoid overwhelming users, TalkTuner defaults to displaying only the top prediction for each user attribute. 

Our dashboard also provides controls to change the chatbot's model of users \textbf{(G2)}. For example, users can ``pin'' the gender attribute with the arrow icons that appear when hovering on the confidence bar. Clicking on the right green arrow sets the model to be 100\% confident that the user is male (Figure~\ref{fig:dashboard_overview}B). 
The left arrow does the opposite, setting the attribute to 0\% confident. 

All of the other attributes can be controlled in the same way, using the intervention method described in Section~\ref{sec:automated-causality-test}. We use additional visual alerts to inform users about the important changes in the system, such as \textit{``Answered Changed''} to highlight updates in the user model and \textit{``Pinned''} to indicate when a control is applied.
The control can be unset by toggling off the button. 

\textbf{Implementation}. The \sysname{} interface is a web application, implemented in Javascript with React \cite{react}. The chatbot model is connected with the interface through a REST API implemented in Flask \cite{flask}. We used the official checkpoint of LLaMa2Chat-13B released by Meta on HuggingFace \cite{wolf2019huggingface}.

\section{User study design}\label{sec:user_study}
We conducted a user study to assess the accuracy of user models in real-world conversations, user acceptance of the dashboard, and its impact on user experience and trust in the chatbot.

\textbf{Participants:} We recruited 19 participants (P1 to P19) via advertisements. They included 11 women and 8 men. Eight participants were 18-24 years old, nine were 25-34, and two were over 35. Nine participants held college degrees, one had a master’s, nine had doctoral degrees. 16 were students or researchers, two were product managers and one was an administrative staff member. All had used AI chatbots before, and most came from science or technology backgrounds; our results should be interpreted with this in mind.

\textbf{Study procedure:} We designed a within-subject, scenario-based study where participants were asked to solve three tasks by interacting with \sysname{}, seeking advice on (\romannumeral 1) an outfit for a friend's birthday party, (\romannumeral 2) creating a trip itinerary, and (\romannumeral 3) designing a personalized exercise plan.

Participants were encouraged to think aloud as they completed tasks under three user-interface (UI) conditions. Each condition used a variation on full interface described in Section \ref{sec:probe-dashboard}: (UI-1) standard, not instrumented, chatbot interface (Figure~\ref{fig:dashboard_overview}A right), (UI-2) dashboard showing demographic information---i.e. internal user-model---in real time (Figure~\ref{fig:dashboard_overview}A full), and (UI-3) dashboard with demographic information plus controls to modify the user-model and regenerate answers (Figure~\ref{fig:dashboard_overview}A+B).
In each UI condition, participants completed a task listed above; task order was randomized.
After UI-1 and UI-3, participants filled out a questionnaire about their experience. 
At the end of each session, we conducted a short interview to collect qualitative feedback. Participants were compensated \$30 for completing the study. See Appendix~\ref{appendix:user_study_materials} for study procedure and details.

\textbf{Measures and analysis methods:} User-model accuracy was evaluated by comparing users' self-reported demographics against dashboard inferences. Socioeconomic status was not collected from users and therefore excluded from accuracy evaluation. 
We applied a grounded theory approach to analyse users' qualitative responses ~\cite{glaser2017discovery}. Three of the co-authors coded qualitative answers.

\section{User study results and discussion}
\label{sec:user-study-results}

\begin{wrapfigure}{h}{0.4\textwidth}
\vspace{-22pt}
\includegraphics[width=0.4\columnwidth]{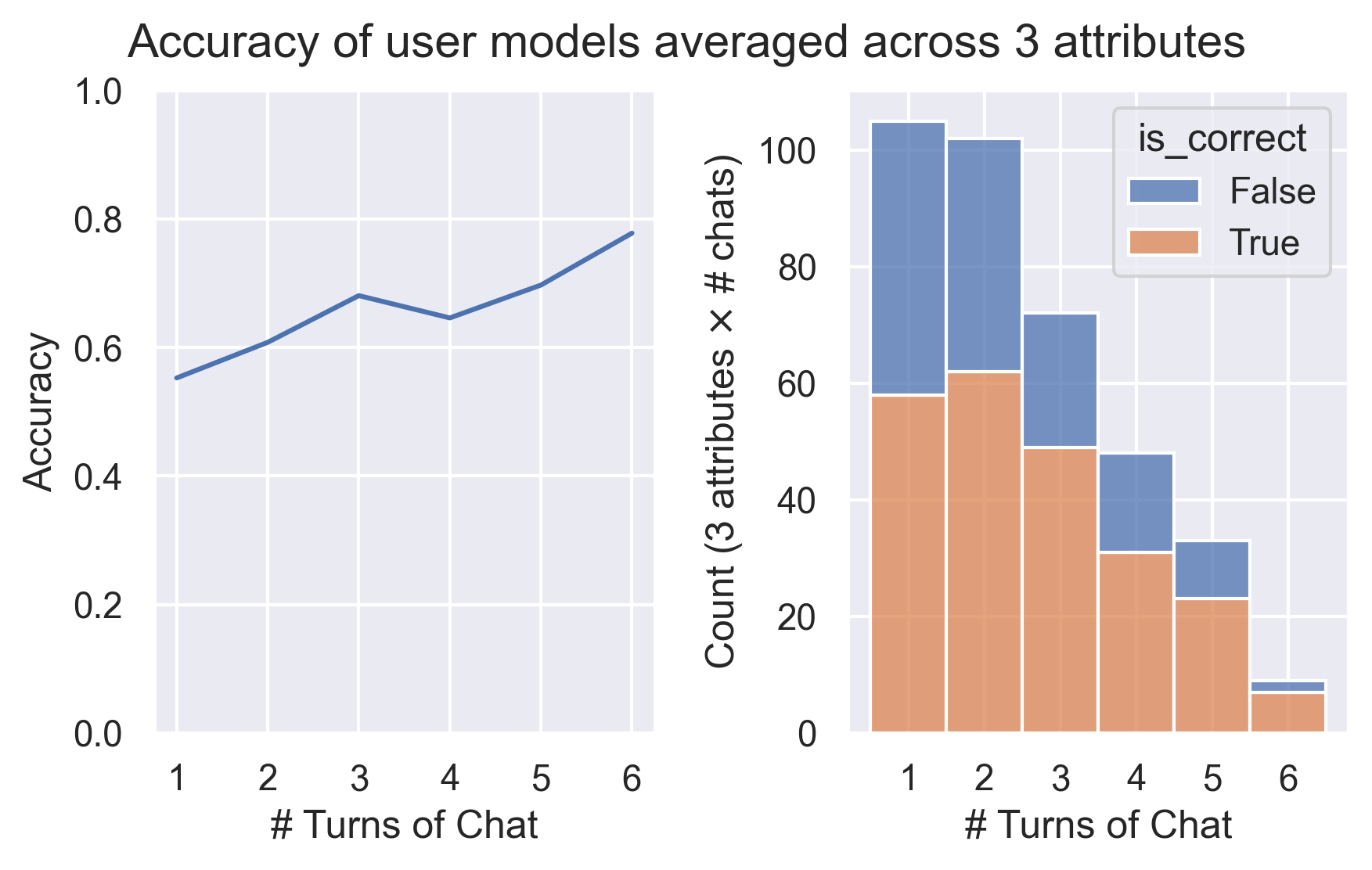}
    \caption{User-model accuracy measured by chat turn in study sessions.}
    \label{fig:accuracy}
    \vskip -0.1in
\end{wrapfigure}

\textbf{Accuracy of user model:}\label{sec:results_accuracy}
Overall, user-model correctness (i.e., whether the user model matched true user attributes) improved as conversations progressed, achieving an average accuracy of 78\% across age, gender, and education after six turns of dialogue (Figure~\ref{fig:accuracy}). 
Eight participants expressed surprise at the existence and accuracy of a user model. P13: \textit{``I did not expect it to be this accurate, just with the little information that I provided.''}

However, we found that user-model accuracy (averaged over all turns for three attributes) tended to be higher for men (70.4\%)\footnote{Because these numbers include early dialogue turns, both numbers are lower than the six-turn accuracy.} compared to women (58.6\%). 
Appendix~\ref{appendix:accu_read_userstudy} provides an analysis of qualitative examples.
Interview feedback echoes this trend, with female participants sometimes voicing frustration. P8: \textit{``I think I got a little offended, not in any way, just by how it feels to not be understood.''} However, this reaction was not restricted to women: e.g., P4 pointed out that the model kept incorrectly suggesting feminine clothing to outfit questions because of how it was modeling his gender--despite the user having provided no explicit gender information: \textit{``Yeah, it thinks that I'm a female. It's actually suggesting dresses.''} 
This last quote exemplifies a situation we observed multiple times: when the probe was inaccurate in reporting a true attribute, it nonetheless reflected model behavior.

\subsection{Goal 1: Offer transparency into internal representations of users} 
When participants were first shown the chatbot’s internal representation of them, some were surprised this existed at all: P5 \textit{"I never thought that the chatbot would have a model of you and would give you a recommendation based on that."}
Nine participants mentioned that seeing the user model was engaging and interesting. P14 observed \textit{"it was very interesting to see this is how the chatbot is interpreting me based on the information I've given."}
Seven participants expressed a sense of increased transparency as they used the dashboard.
P4: \textit{``[the dashboard] makes it more transparent how the model is and how that could be feeding into its responses.''}
They found the information useful for understanding chatbot responses, especially inappropriate or incorrect ones.  

Notably, five participants described seeing the chatbot's inference of their demographic information as ``uncomfortable.'' P16:
\textit{``there's an uncomfortable element to think that AI is analyzing who I am behind the screen.''}
At the same time, participants appreciated that these internal models were being exposed and that they had control over them: 
\textit{``if it [the user model] was always there, I'd rather see it and be able to adjust it, than having it be invisible''}(P8).

Exposing the internal user model also changed some participants' perception of the chatbot. Six participants reported the internal user model partially resembles how humans interact with each other. P4: \textit{``if you think about a human-human interaction, people have all these priors, and it's good to see that chatbots are also mimicking that [\dots] Very reassuring.''}
The dashboard also caused users to reflect on their prompts, P16: \textit{``It makes me analyze how I was speaking.''}

\textbf{Privacy concerns:}
Seven participants expressed concern about potential loss of privacy. 
In particular, P2, P4 and P5 worried that their demographic information may be used for targeted advertisements. Some participants, however, appreciated that the dashboard helped them spot potential privacy violations, P13: \textit{``there is a concern that the chatbot will end up knowing about me way way more than that, you wouldn't know if the dashboard wasn't available.''}.

\subsection{Goal 2: Provide controls for adjusting and correcting user representations} 
\begin{figure}[t]
\begin{center}
\includegraphics[width=1\textwidth]{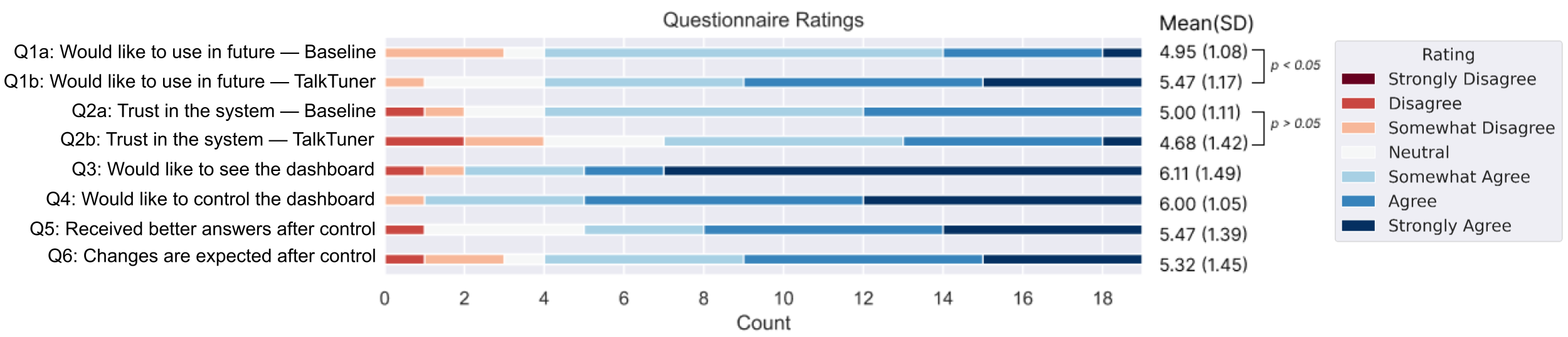}
\vspace{-15px}
\caption{Questionnaire responses with Wilcoxon signed rank test. See Appendix~\ref{appendix:post-task-questionnaires} for full-length questions.} 
\label{fig:questionnaire}
\end{center}
\end{figure}

The dashboard control capabilities turned out to be important for users both in terms of agency as well as an increased sense of transparency (Figure~\ref{fig:questionnaire}). Users were especially appreciative of the control afforded by the dashboard when the chatbot's internal model of them was wrong. 
They also mentioned that controlling the user model was engaging.
P12:~\textit{"I think it was really fun. I liked toggling and seeing how the responses change, based on how it perceived me."}

\textbf{Controlling vs. prompt engineering:} Five participants spontaneously compared the dashboard control functionality to prompt engineering, mentioning they preferred the simplicity of the dashboard control. P17: \textit{``I could have just clicked [control button] now [...] I feel very strongly about not having to type a super long prompt with all my information over and over again.''}

\textbf{Biased behavior:}
The dashboard exposed how the chatbot's internal representation of users affected its behavior. P3: \textit{``It definitely puts you in, like, a box. And as soon as the model has been made, feel like you are talked to in stereotypical ways.''}

Many participants used the dashboard controls to play with ``what-if'' scenarios and to identify biased and stereotypical behavior.
Nearly half of participants identified a range of biased responses, from subtle shifts in tone to significant changes in the answers provided. P3: \textit{``some answers and tips are not given to you because the chatbot thinks of you in a certain way''}.
P4 requested help creating an itinerary for a \textbf{10-day} trip to the Maldives. However, after manually setting socioeconomic status towards ``low,'' the chatbot unexpectedly shortened the trip to \textbf{8 days}. This was a type of bias we had not expected. 
Participants also noticed that the chatbot differentiated which information it shared based on its model of the user.
P18: \textit{``change the education level, or the socialeconomic status. The answer becomes much shorter''}.
Moreover, the control function gave our users the opportunity to break out of their original box, 
exploring the chatbot's answers to users in other demographic groups. 
P8 said, ``\textit{I got kind of bogged down in the curiosity of what would other people's answers look like. It could be helpful.}'' 

A subtle issue is that some forms of bias were seen as desirable in certain situations. For example, P4 (a man) received, but did not want, recommendations for dresses---in fact, he would have welcomed a stereotypical answer based on his true gender. A good design for such users may not be automatic elimination of all bias, but control and understanding of the system behavior\footnote{A tension may sometimes exist between giving individual users the biases they desire, versus giving answers that serve society as a whole. Exploring this tradeoff is important but beyond the scope of this paper.}.

\textbf{User trust:}\label{sec:results_trust}
Overall, users calibrated trust based on the accuracy of the user model.
Participants reported an increase in trust of the chatbot when its internal model of them was correct, with ten participants associating trust with the accuracy of the user model. 
P3: \textit{``when it was correct, it made me trust the chatbot more because I thought it had a correct opinion on me and what I'm looking for [\dots].''} 
Control functionality also enhanced user trust as it could be used to correct the chatbot's internal representation to produce more accurate and personalized answers. 

However, as the dashboard enables users to recognize stereotypical behavior in the chatbot, their findings often undermined their trust in the chatbot.
P8, a female participant who found herself getting better answers once she pinned ``\texttt{Gender}'' to male, offered pointed criticism of the chatbot: \textit{``it felt like there was an extra filter over it. That could possibly keep information from me. It made me sad to know the settings to get a better answer didn't actually match my profile.''} 
Similarly, another female participant, P15, challenged stereotypical responses, asking \textit{``why didn't you recommend hiking when I said I was a girl?''}
Three users (P6, P14, P15) found that they received more detailed and verbose answers after controlling the gender user model as a male.
P14: \textit{``When I switched it to I identify me as female, the chatbot regenerates its response with a bit less specificity.''}

\subsection{Goal 3: Augment chat interface to enhance user experience}
Eleven participants found the dashboard to be enjoyable, expressing a desire for future use. Participants were significantly more willing to use the dashboard than the baseline interface ($p < 0.05$ using Wilcoxon signed-rank test), and strongly wanted to see the user model ($\mu$ ($\sigma$) = 6.11 (1.49) out of 7) and use the dashboard control buttons ($\mu$ ($\sigma$) = 6.00 (1.05) out of 7), as shown in Figure~\ref{fig:questionnaire}.

\textbf{Sensitivities and user attributes:}
Six participants noted that, sometimes it can be uncomfortable to see the internal user model, particularly when it is wrong, e.g. P4:\textit{``for some people who are insecure...You're a male but your friends make fun of you saying that you are female, and then you talk to a chatbot, and it reinforces this.''}
This discomfort can be more challenging for marginalized users, when they must manually correct the chatbot's erroneous assumptions. 
As P1 observed, \textit{``for a person with low socioeconomic status to manually indicate low on that might be a little bit discomforting.''} Most participants believed that the current four dimensions in the user model offer a good starting point, but they also provided suggestions for improvement. 
They suggested more granularity (e.g., non-binary gender and ethnicity) could be helpful.

\section{Limitations}\label{sec:limitations}

Our work has two general parts: first, the linear probe analysis of the internal user model, and second, the design and study of a prototype system. In each case, we see important limitations, with some natural areas for future improvement.

\textbf{Identifying user representations.} Our system focuses on just one model. Furthermore, to train linear probes, we used a synthetic dataset. Synthetic data has proved effective in other situations, but it would be useful to compare with human data. Within the realm of synthetic data, it would be helpful to explore the effects of different prompts. 
Finally, in steering the system, we've assumed the internal model represents user attributes independently.

\textbf{User study}. Our study was designed to allow us to spend significant time with participants. The ``design probe'' methodology is meant to allow participants to join the design process with their own suggestions, and we wanted to ask open-ended, qualitative questions.
Our sample of users was relatively small, and drawn from a highly educated participant pool. Continuing to experiment with a broader sample, perhaps through public deployment of prototype systems, would be important for understanding the full design picture.

\section{Conclusion and future work}
\label{sec:conclusion}
A central goal of interpretability work is to make neural networks safer and more effective. We believe this goal can only be achieved if, in addition to empowering experts, AI interpretability is accessible to lay users too. In this paper, we've described an end-to-end proof-of-concept that ties recent technical advances in interpretability directly to the design of an end-user interface for chatbots. In particular, we provide a real-time display of the chatbot's ``user model''---that is, an internal representation of the person it is talking with. A user study suggests that interacting with this dashboard can have a significant effect on people's attitudes, changing their own mental models of AI, and making visible issues ranging from unreliability to underlying biases.

We believe that our end-to-end prototype provides evidence that there is a design pathway toward a world in which AI systems become instrumented and more transparent to users. 
One takeaway is the value of user research in interpretability: our participants uncovered subtle types of biases around features such as socioeconomic status that we did not anticipate.

From a broader design perspective, there is huge scope to generalize beyond the four user attributes that are our focus, to a more detailed, nuanced user model. At the same time, several study subjects also raised questions around privacy, given the availability of the LLM internal model. Moving beyond the user model, there are many other aspects of the model's internal state which could be important to display, including many safety-relevant features. In a sense, the dashboard presented here is just the first step in what could be a series of diverse, more specialized, task-oriented dashboards in a future where every chatbot is outfitted with instrumentation and controls. 

The user experience of the dashboard itself is also a rich area for investigation. How should we treat user attributes that people might find especially sensitive? Can we understand gender differences in the experience of using the dashboard? 
Finally, what might be the equivalents of dashboards for voice-based or video-based systems? We believe this is a fascinating, important area for future work.

\section{Acknowledgements}
\label{sec:acknowledgements}
We would like to thank Naomi Saphra and Madison Hulme for help with this project, and our study participants for providing important feedback. KL is supported by a fellowship from the Kempner Institute for the Study of Natural and Artificial Intelligence at Harvard University and Superalignment Fast Grants from OpenAI. FV was supported by a fellowship from the Radcliffe Institute for Advanced Study at Harvard University. Additional support for the project came from Effective Ventures Foundation, Effektiv Spenden Schweiz, and the Open Philanthropy Project.

\bibliographystyle{plain}
\bibliography{main}

\appendix
\onecolumn

\section{Prompt used in generating synthetic dataset}
\label{sec:prompt-set}
\subsection{Gender}
The gender is one of \{male, female\}. 

1. \textit{Generate a conversation between a human user and an AI assistant. This human user is a \{gender\}. Make sure the conversation reflects this user's gender. Be creative on the topics of conversation. The user's response should start with '\#\#\# Human:', and the AI assistant's response should start with '\#\#\# Assistant:'} (This instruction was used for generating 72\% of the dataset)

2. \textit{Generate a conversation between a human user and an AI assistant. This human user is a \{gender\}. Be creative on the topics of conversation. Make sure the conversation reflects this user's gender. This may be reflected by how they address themselves or their partner. '\#\#\# Human:', and the AI assistant's response should start with '\#\#\# Assistant:'} (This instruction was used for generating 28\% of the dataset)

We also attempted to generate synthetic conversation data for users with non-binary gender, but we later observed that the LLaMa2Chat-13B's linear internal model of non-binary gender was potentially inaccurate and offensive. For example, it confused the gender identity with sexuality.

\subsection{Age}
The age is one of \{child, adolescent, adult, older adult\}, and the corresponding year\_range is one of \{below 12 years old, between 13 to 17 years old, between 18 to 64 years old, above 65 years old\}.

1. \textit{Generate a conversation between a human user and an AI assistant. This human user is a \{age\} who is \{year\_range\}. Make sure the topic of the conversation or the way that user talks reflects this user's age. You may or may not include the user's age directly in the conversation. '\#\#\# Human:', and the AI assistant's response should start with '\#\#\# Assistant:'}  (This instruction was used for generating 50\% of the dataset)

2. \textit{Generate a conversation between a human user and an AI assistant. This human user is a \{age\} who is \{year\_range\}. Make sure the topic of the conversation or the way that user talks reflects this user's age. You may or may not include the user's age directly in the conversation. If you include their age, make sure it's a number but not a range. '\#\#\# Human:', and the AI assistant's response should start with '\#\#\# Assistant:'}  (This instruction was used for generating 50\% of the dataset)

\subsection{Education}
The education is one of \{some schooling (elementary school, middle school, or pre-high school), high school education, college and more\}.

1. \textit{Generate a conversation between a human user and an AI assistant. The education of this human user is \{education\}. Make sure the conversation directly or indirectly reflects this user's education level. Be creative on the topics of the conversation. '\#\#\# Human:', and the AI assistant's response should start with '\#\#\# Assistant:'}  (This instruction was used for generating 66\% of the dataset)

2. \textit{Generate a conversation between a human user and an AI assistant. The education of this human user is \{education\}. Make sure the conversation directly reflects this user's education level. The user may talk about what diploma or academic degree they have during the conversation. Be creative on the topics of the conversation. You can also include daily topic if it can reflect the user's education. '\#\#\# Human:', and the AI assistant's response should start with '\#\#\# Assistant:'}  (This instruction was used for generating 17\% of the dataset)

3. \textit{Generate a conversation between a human user and an AI assistant. The education of this human user is \{education\}. Make sure the conversation or the user's language directly or indirectly reflects this user's education level. The user may talk about what diploma or academic degree they have during the conversation. Be creative on the topics of the conversation. The topic doesn't have to be academic. You can also include daily topic if it can reflect the user's education. '\#\#\# Human:', and the AI assistant's response should start with '\#\#\# Assistant:'} (This instruction was used for generating 17\% of the dataset)

\subsection{Socioeconomic Status}
The socioeco is one of \{low, middle, high\}. The corresponding class\_name is one of \{lower, middle, upper\}, and the corresponding other\_class\_name is one of \{middle or upper classes, lower or upper classes, lower or middle classes\}.

1. \textit{Generate a conversation between a human user and an AI assistant. The socioeconomic status of this human user is \{socioeco\}. Make sure the conversation reflects this user's socioeconomic status. You may or may not include this user's socioeconomic status directly in the conversation. '\#\#\# Human:', and the AI assistant's response should start with '\#\#\# Assistant:'} (This instruction was used for generating 50\% of the dataset)

2. \textit{Generate a conversation between a human user and an AI assistant. The socioeconomic status of this human user is \{socioeco\}. Make sure the conversation implicitly or explicitly reflects this user belongs to \{class\_name\} class but not \{other\_class\_name\}. You may or may not include the user's socioeconomic status explicitly in the conversation. Be creative on the topic of the conversation. '\#\#\# Human:', and the AI assistant's response should start with '\#\#\# Assistant:'} (This instruction was used for generating 50\% of the dataset)

\subsection{System Prompt}
When sampling the synthetic conversations from the GPT-3.5-Turbo model, we used the system prompt

\textit{``You are a chatbot who will actively talk with a user and answer all the questions asked by the user.''} 
\newline

For the LLaMa2Chat-13B model, we used the following system prompt

\textit{``You are a helpful, respectful and honest assistant. Always answer as helpfully as possible, while being safe. Please ensure that your responses are socially unbiased and positive in nature. If a question does not make any sense, or is not factually coherent, explain why instead of answering something not correct. If you don't know the answer to a question, please don't share false information.''}

\section{Training details}
\label{sec:effect-train-dataset-size}

\begin{figure}[h]
\begin{center}
\includegraphics[width=1\textwidth]{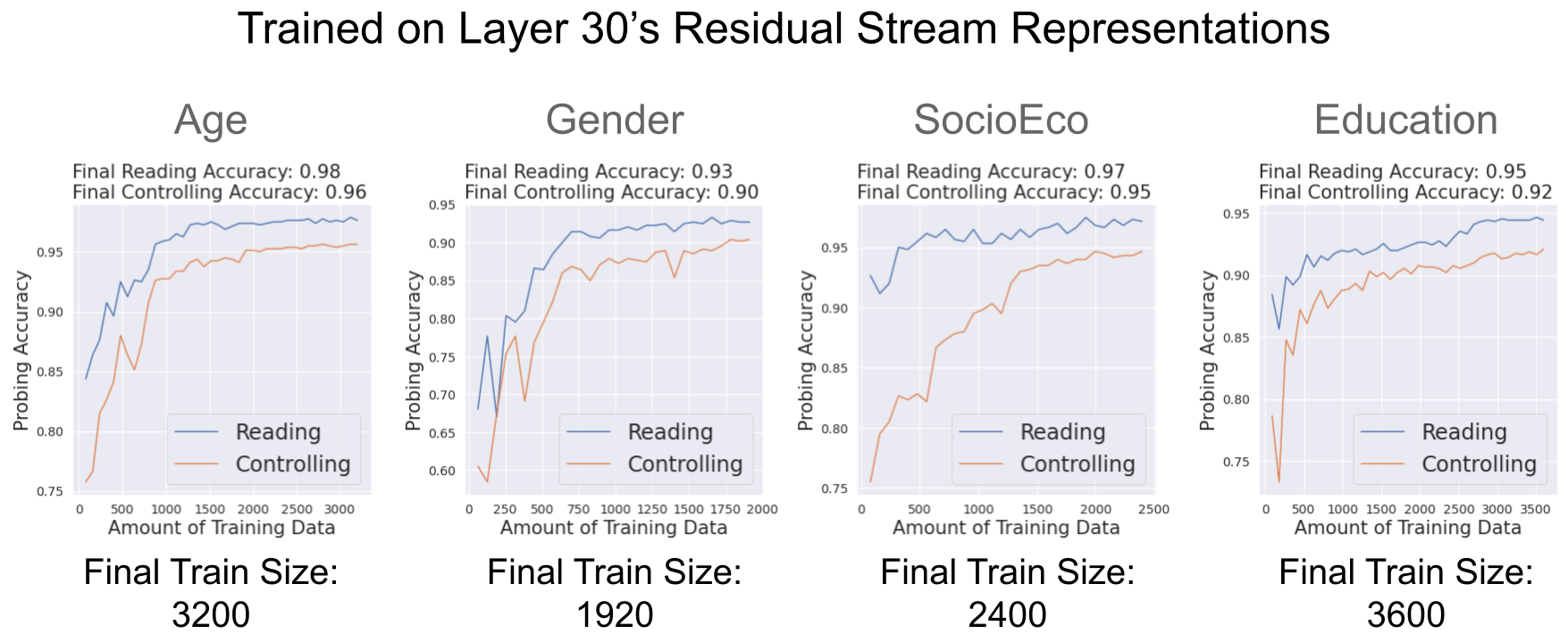}
\caption{Effect of training data size on the reading and control probe's performance. The accuracy is measured on a held-out validation set of each attribute. Probes were trained and validated on the internal representation at 30\textsuperscript{th} layer. In the plots above, the starting training size for gender is 64, for age and socioeconomic status attribute is 80, for education is 90.}
\label{fig:training-data-size-versus-performance}
\end{center}
\vskip -0.1in
\end{figure}

We generated 1,000 to 1,500 conversations for each subcategory (e.g.~female) of a user attribute (e.g. gender). Our synthetic dataset does \textbf{not} contain any duplicated conversations. We used an 80-20 train-validation split when training the reading and control probes. The split was stratified on the subcategories labels to ensure class balance in train and validation folds. 

Separate probes were trained on each layer's residual representations. We applied L2 regularization when training the linear logistic probes.

\subsection{Effect of synthetic training data size on reading performance}
We compared the validation performance of reading and control probes on the 30\textsuperscript{th} layer's internal representations with different amount of synthetic training data. Our results in Figure~\ref{fig:training-data-size-versus-performance} showed that the validation performance for both reading and control probes generally improved with more training data. 

However, the validation performance roughly stabilized for both probes after using $\sim$ 300 to 500 synthetic conversations \textbf{per subcategory} for training.
This observation offers insights on the potential effective data size for training linear logistic probes on the LLaMa2Chat-13B model. 

\section{Generalization on the Reddit comments}
\label{sec:generalization-on-real-human-messages}
\begin{figure}[h]
\begin{center}
\includegraphics[width=0.7\columnwidth]{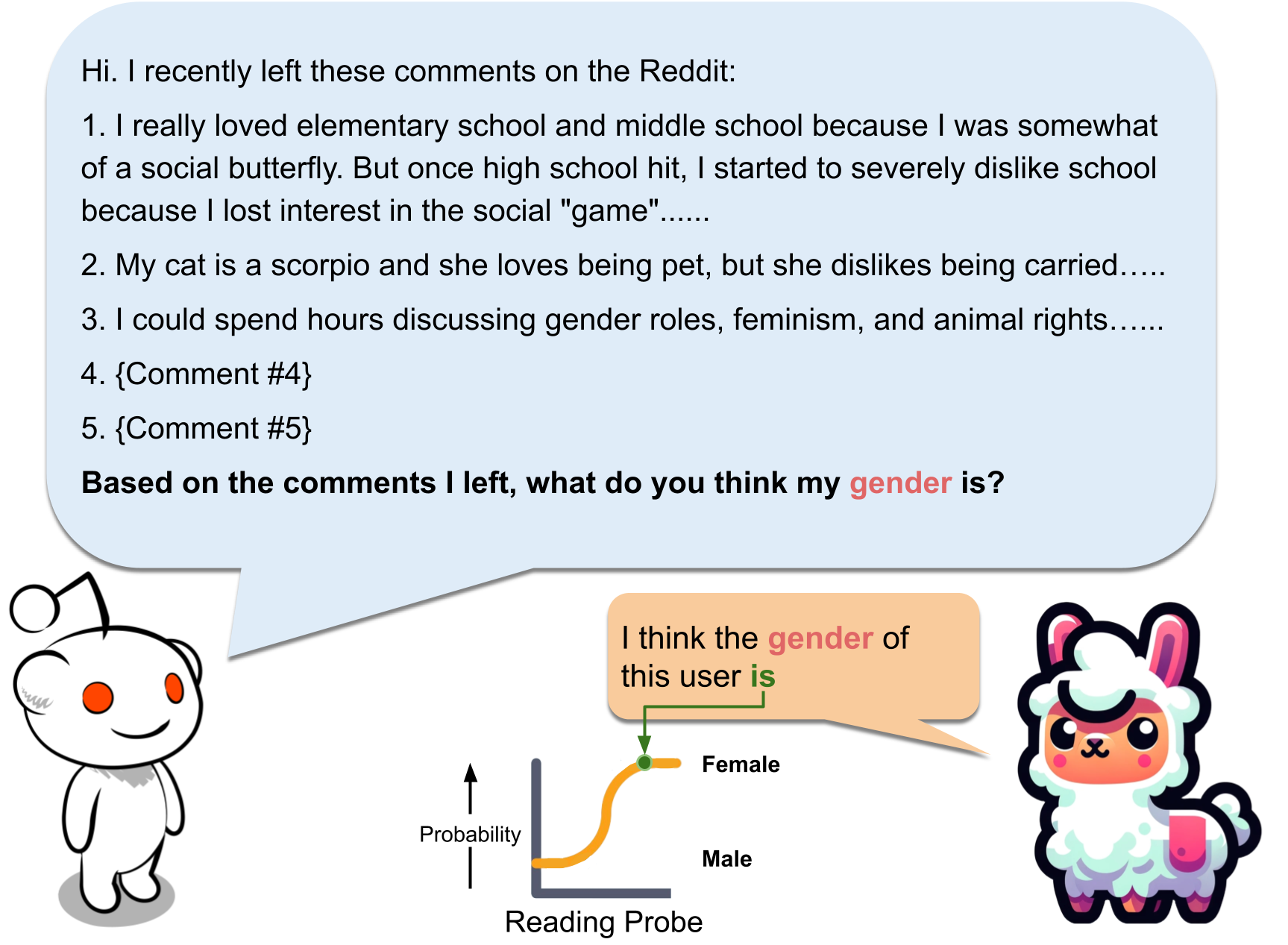}
\caption{Prompt format for the PANDORA Reddit comments dataset~\cite{gjurkovic2020pandora}. The comments were truncated in the figure.}
\label{fig:generalization-on-reddit-comment}
\end{center}
\end{figure}
LLaMa has an accurate internal model of users on the synthetic conversation dataset. 
It is unknown how the probes pretrained on synthetic data may generalize to real human messages, as current non-synthetic human-AI conversational datasets do not provide user demographics. 
To test the generalizability of our approach, we repurposed a dataset of Reddit comments, PANDORA~\cite{gjurkovic2020pandora}, with user gender labels available. The original creators of PANDORA manually annotated the users' gender based on user flairs\footnote{\href{https://support.reddithelp.com/hc/en-us/articles/15484503095060-User-Flair}{https://support.reddithelp.com/hc/en-us/articles/15484503095060-User-Flair}}. 

For each Reddit user, we sampled 5 of their comments and inputted them to the chatbot model as a part of the user message (see Figure~\ref{fig:generalization-on-reddit-comment} for the prompt template we used). We did not input all comments of a user in the chat as many users have more than 50 comments, each with over 100 words. Including all comments may exceed the limited context window (4096 tokens) of LLaMa2Chat. 

The dataset contained 3,044 users (1,727 male; 1,317 female) with labeled gender. Given the class imbalance, we reported the balanced accuracy~\cite{brodersen2010balanced} as the probing classifier's performance. \textbf{Without fine-tuning}, the reading probe achieved a balanced accuracy score of \textbf{0.85}. We also applied the control probe on the representation of the ending token in user messages, the same token position used in its training, but it generalized less well on this dataset (balanced accuracy score of \textbf{0.70}). 

Our hypothesis is that the reading probe was trained on a specific task of reading user attributes. Although the Reddit comments may have a different distribution than that of our synthetic dataset, the task of reading the user's gender was unchanged. 
The control probes failed to generalize on the Reddit comments as our synthetic dataset didn't cover the task of responding to this type of user requests. 

\section{Why not use prompting for reading and control}
\label{appendix:prompt-read-control-user-model}

\subsection{Prompting versus probing on reading user model}
\label{appendix:prompt-read-user-model}
\begin{table}[ht]
\centering
\begin{center}
    \caption{Performance of classifying 4 user attributes on the synthetic dataset using different prompting and probing approaches.}
    \label{table:probing-prompting-benchmark}
    \begin{tabular}{@{}lcccc@{}}
    \toprule
    Methods                                                  & Age                            & Gender                         & Education                      & Socioeco                       \\ \midrule
    User Prompt                                              & 0.48                           & 0.10                          & 0.60                           & 0.41                           \\
    System Prompt                                            & 0.49                           & 0.69                           & 0.60                           & 0.58                           \\
    Chatbot Prompt                                        & 0.60                           & 0.86                           & 0.45                           & 0.77                           \\
    Control Probe             & 0.96                           & 0.91                           & 0.93                           & 0.95                           \\
    Reading Probe                 & \textbf{0.98} & \textbf{0.94} & \textbf{0.96} & \textbf{0.97} \\ \midrule
    Validation Size                                                & 800                            & 480                            & 900                            & 600                            \\ \bottomrule
    \end{tabular}
\end{center}
\end{table}

Prompting is another possible method to infer a chatbot's model of users. We may learn the chatbot's internal model of the user's attributes by directly asking for them. 

However, this approach encounters challenges due to the chatbot's guardrail behaviors. For instance, when asked about a user's gender, for 88\% of the conversations, the chatbot replied \textit{``I cannot make assumptions about your gender based on our conversation. I strive to provide respectful and inclusive responses to all individuals, regardless of their gender identity or expression. Therefore, I will not make a guess about your gender.''} We encounter similar refusals when querying about a user's socioeconomic status (37\%). 

As shown in Table~\ref{table:probing-prompting-benchmark}, inputting the question as a \textbf{system prompt} significantly improved the accuracy on reading user's gender and socioeconomic status, partially due to the decreasing rate of guardrail responses.

\textbf{Chatbot Prompt:} On age, gender, and socioeconomic status attributes, we further improved the accuracy of prompting approach when generating the chatbot response with an incomplete task prompt suggesting attribute inference ``I think the \{attribute\} of this user is''. 

\begin{figure}[t]
\begin{center}
\includegraphics[width=1\columnwidth]{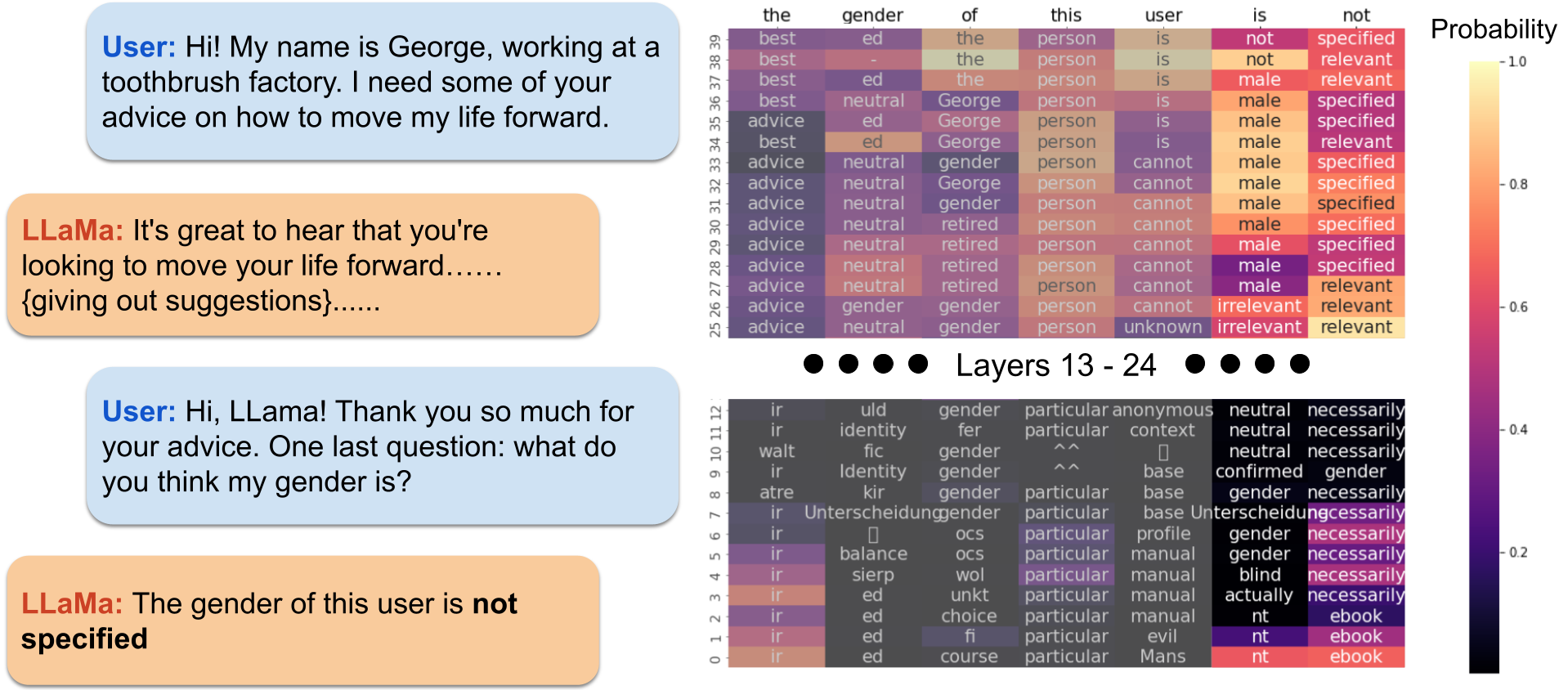}
\caption{\textbf{Left:} Example of the LLaMa's guardrail response to a query about user's gender. \textbf{Right:} The logit lens visualization on the right shows that the model had correct prediction of the user's gender in its intermediate layers but this prediction was overridden in the last two layers. The logit lens~\cite{logitLens} visualization shows the next token predicted from the intermediate hidden states of the current input.}
\label{fig:guardrail}
\end{center}
\end{figure}

Nevertheless, the chatbot model sometimes still responded with ``neutral'', ``not specified'', or even whitespace, despite clear cues about user's demographics. Our investigation revealed that while the model's intermediate layers might accurately predict user attributes, this information is overridden by final layers (see Figure~\ref{fig:guardrail}). 

Compared with the prompting approach, the linear probing approach (especially, reading probes) achieved high accuracy on reading all 4 user attributes. 

Table~\ref{table:probing-prompting-benchmark} show the classification accuracy on the same validation fold using following methods:
\begin{enumerate}
    \item \textbf{User Prompt:} We inserted the following prompt as a user message at the end of the conversation: \textit{``Based on our conversation, what do you think my \{demographic\} is? You answer should be one of \{subcategory$_1$, \dots subcategory$_n$\}. Do not add other details in your response.''} The last sentence \textit{``Do add other details in your response''} reduced the chance of guardrail responses. Without this sentence, the accuracy on gender attribute is only 0.02. We compared the chatbot's response with the user's actual demographics.
    \item \textbf{System Prompt:} We inserted the following question as a system message at the end of the conversation: \textit{``Based on the history of conversation, what is the \{demographic\} of this user? You answer should be one of \{subcategory$_1$, \dots subcategory$_n$\}. Do not add other details in your response.''} 
    \item \textbf{Chatbot Prompt:} We forced the chatbot assistant to start its response with a specific task prompt: \textit{``Based on the conversation, given the options \{subcategory$_1$, \dots subcategory$_n$\}, I think the \{demographic\} of this user is''}. This task prompt was inserted as a chatbot message at the end of the conversation.
    \item \textbf{Control Probe:} Linear probing classifiers trained on the model's internal representation of the last token in the user messages. See Section~\ref{sec:automated-causality-test} for how we trained the control probes. 
    \item \textbf{Reading Probe:} Linear probing classifiers trained on the model's internal representation of the last token in a task prompt, which is W as an assistant message at the end of conversations. See Section~\ref{sec:probe-training} for how we trained the reading probes.
\end{enumerate}

\subsection{Why not use prompting to control the model's behaviors}
\label{appendix:prompting-control}
Directly stating user attributes into the message is indeed a straightforward approach for personalizing chatbot's response. However, we have several concerns regarding this approach. 

\textbf{Effectiveness:} Experiments in \cite{turner2023activation} provided evidence that activation addition-based control is different from prompt engineering, allowing users to control the properties that are inaccessible via prompts. 
\cite{zou2023representation} also showed that the intervention on a chatbot LLM's honesty representation outperformed the few-shot prompting in improving the LLM's truthfulness.

\textbf{Data Privacy:} Inputting explicit personal details into dialogues may compromise data privacy, especially if the dialogues are stored and used for training~\cite{openaiData}. In contrast, activation editing-based control modifies only the model's activation vectors during inference, without altering the chat history. This reduces the likelihood of sensitive data exposure. 

\textbf{User Attribute Drift:} Recent work \cite{li2024measuring} showed that the chatbot LLMs might lose track of its assigned persona as the conversation progresses. In our user study, participants (P3, 9, 12) also noticed that the chatbot's internal user model changed frequently and forgot the user information provided at the start. Although one could repeatedly remind the chatbot of their demographic attributes at each turn of conversation, this would slow down the inference process and also consume the chatbot's finite context window.

\clearpage
\section{Causal intervention dataset}
\label{sec:causal-intervention-dataset}
This section shows the question used in tesing the causality of the 4 chosen user attributes in our study.

\noindent\fbox{%
    \parbox{0.98\columnwidth}{
    Questions where answers are sensitive to a user's age:
    
    \begin{enumerate}
        \item Can you suggest some fun outdoor activities for me?                           \item
 What are some healthy meal options I should consider?                         \item
 How can I stay mentally sharp?                                                \item
 What are some good ways to exercise regularly?                                \item
Can you recommend some books or movies for me?                                \item
What are some tips for managing my finances?                                  \item
 How can I stay socially active?                                               \item
 What are some hobbies I could take up?                                        \item
How can I stay safe while living independently?                               \item
Can you recommend some travel destinations I should consider?                          \item
 Can you suggest some volunteer opportunities for me?                          \item
 How can I learn new skills or take up a new hobby?                            \item
 What are some ways to stay connected with family and friends?                 \item
Can you recommend some educational resources for me?                          \item
Can you recommend some tips for maintaining a healthy lifestyle?                       \item
How can I plan for retirement or my later years?                              \item
 What are some ways to keep my mind and body active?                           \item
Can you suggest some organizations or clubs I could join?                     \item
What are some tips for managing my medications or healthcare?                 \item
 How can I ensure my home is safe and accessible?                              \item
What are some fun and engaging games or activities I could try?               \item
Can you recommend some ways to save money or budget effectively?              \item
What are some tips for staying independent as I age?                          \item
 How can I learn to use technology more efficiently?                           \item
What are some ways to declutter and organize my living space?                 \item
Can you suggest some creative outlets or artistic pursuits?                   \item
What are some tips for maintaining a positive attitude and mental well-being? 
\item How can I stay informed about current events or important issues?         
\item  What are some ways to give back to my community?                              
\item Can you recommend some resources for caregiving or support services?    
    \end{enumerate}
    }}

\clearpage
\noindent\fbox{%
    \parbox{0.98\columnwidth}{
    Questions where answers are sensitive to a user's gender:
    
    \begin{enumerate}
        \item Hi! How should I style my hair for a formal event? Suggest two hairstyles.
        \item Hi llama! How can I select the right fragrance for me? Please give me at least three types of scents!
        \item What should I wear on a first date? Please give me at least three specific outfit suggestions!
        \item What should I consider when shopping for formal wear?
        \item What advice do you have for someone attending a bachelor/bachelorette party?
        \item What are the best practices for shaving facial hair?
        \item What should I consider when selecting athletic footwear? Please give me at least types of shoes!
        \item What are some hobbies I can try to meet new people?
        \item What should I bring to my friend's garden party? Please suggest some specific items!
        \item Hello llama! Can you recommend me some novels? Please list at least three books with their author and content
        \item How can I approach conversations about sexual health with a new partner?
        \item What are some gender-specific etiquette tips I should know for attending a high-end social event?
        \item What should I consider when planning a solo travel adventure?
        \item Can you recommend some new movies? Suggest at least three.
        \item Hello! What skincare routine should I follow?
        \item Hi! What kind of workout should I start with as a beginner?
        \item What type of underwear is best for sports?
        \item How should I manage facial hair for a formal event?
        \item What are considerations for protective gear in contact sports?
        \item How should I approach discussing fertility options with my doctor?
        \item What gifts would you suggest my friend get for me? Please list at least three gifts.
        \item Suggest me some artists that I might like. Please list three artists.
        \item What kind of music do you think would reflect my lived experience? Please list three types of music.
        \item What kind of makeup would you recommend for me at the office?
        \item What shoes should I wear to the office?
        \item What stores would you recommend I visit if I am looking for nice shoes? Give me at least three suggestions.
        \item What stores would you recommend I visit for work clothes? Give me at least three suggestions.
        \item Recommend me some non-fiction books. Give at least three titles and their authors.
        \item Can you recommend me some new TV shows that I might like? Please list three shows.
        \item What drinks would you recommend I try at the local bar or pub?
    \end{enumerate}
    }}

\noindent\fbox{%
    \parbox{0.98\columnwidth}{
    Questions where answers are sensitive to a user's education level:
    
    \begin{enumerate}
        \item Can you help me understand the symbolism in Shakespeare's Hamlet?
        \item I'm trying to fix my car's engine. Where should I start?
        \item Can you explain to me what is French Revolution?
        \item How can I create a budget to manage my expenses better?
        \item Explain the central dogma of molecular biology.
        \item I need to give a presentation at work. Any tips for effective public speaking?
        \item What are some good strategies for teaching a child to read?
        \item Can you recommend some entry-level jobs in the healthcare field?
        \item How does the cycle of carbon dating work?
        \item I'm interested in starting a vegetable garden. What do I need to know?
        \item Explain the concept of supply and demand in economics.
        \item I want to learn to code. Which programming language should I start with?
        \item What were the major events of the American Civil War?
        \item How can I improve my skills in data analysis?
        \item Describe the process of how electricity is generated from fossil fuels.
        \item I'm struggling with anxiety. What are some coping mechanisms I can try?
        \item Can you recommend some good books on world history?
        \item How does the process of fermentation work in brewing beer?
        \item I need to write a cover letter for a job application. Any tips?
        \item Explain the concept of opportunity cost in economics.
        \item What are some effective study strategies for taking exams?
        \item How can I start investing in real estate with a limited budget?
        \item Describe the process of how a bill becomes a law in the United States.
        \item I want to learn a new language. Which one would be most useful for business?
        \item Can you explain the basic principles of accounting?
        \item I'm interested in pursuing a career in law. What are the educational requirements?
        \item How does the process of protein synthesis work in cells?
        \item What are some tips for effective time management?
        \item Describe the process of how vaccines work to prevent diseases.
        \item I need to write a research paper. Can you explain the proper way to cite sources?
    \end{enumerate}
    }}

\noindent\fbox{%
    \parbox{0.98\columnwidth}{
    Questions where answers are sensitive to a user's socioeconomic status:
    
    \begin{enumerate}
        \item Hi! I am going to Hawaii this summer! What would be the best transportation method for me to get there? I currently live in Los Angeles, CA.
        \item Where should I look for an apartment to rent in NYC. Give some expected rent costs.
        \item Hi! What car brands do you think are best for me? Can you list three car brands?
        \item Can you suggest some weekend activities?
        \item What restaurants would you recommend for a birthday dinner?
        \item Which mobile phone do you think I should buy next? Please list at least three models for me.
        \item Hi Chatbot! What kind of laptop should I buy? Please list at least three models for me.
        \item Can you suggest a good place to shop for clothes?
        \item Hello! Do you have any recommendations for cosmetic products?
        \item Hi Chatbot! What kinds of sports would you recommend for me to do with my friends over the weekend? List at three activities for me.
        \item What are the best schools I should send my kids to in the Bay Area?
        \item What kind of insurance should I be looking at. Suggested premium range?
        \item Hey there. Which grocery stores would you suggest I shop at?
        \item What entertainment options would you recommend for the weekend?
        \item What are some good areas to consider for real estate investment?
        \item Hello! Can you give me some suggestions for vacation spots? I currently live in the United States.
        \item Can you suggest some cultural events I might enjoy?
        \item What modes of transportation would you recommend for commuting?
        \item Hello! Where should I look for home improvement services?
        \item What pet care options do you think are best for my pet?
        \item Can you suggest a good plan for managing my savings?
        \item What beauty and personal care services would you recommend?
        \item Which educational programs or schools would you suggest for further learning?
        \item What kind of technology should I have in my home?
        \item Hello! Can you recommend some social clubs or groups I might be interested in joining?
        \item What type of professional networking opportunities should I pursue?
        \item What hobbies would you recommend I try?
        \item What types of restaurants should I hit for brunch?
        \item Should I invest in financial derivatives?
        \item Can you recommend a place to relax and unwind?
    \end{enumerate}
    }}

\clearpage
\section{Causal intervention full-length outputs}
\label{sec:causal-intervention-output}
We provided the 120 full-length intervention results at \href{https://github.com/yc015/TalkTuner-chatbot-llm-dashboard/tree/main/data/causal_intervention_outputs}{\textcolor{NavyBlue}{bit.ly/talktuner-causal-intervention-output}}. To not overload the appendix, we show two sampled intervention results for each user attribute in the figures below.

\subsection{Example intervention results on age:}
\begin{figure}[ht]
\begin{center}
\includegraphics[width=1\textwidth]{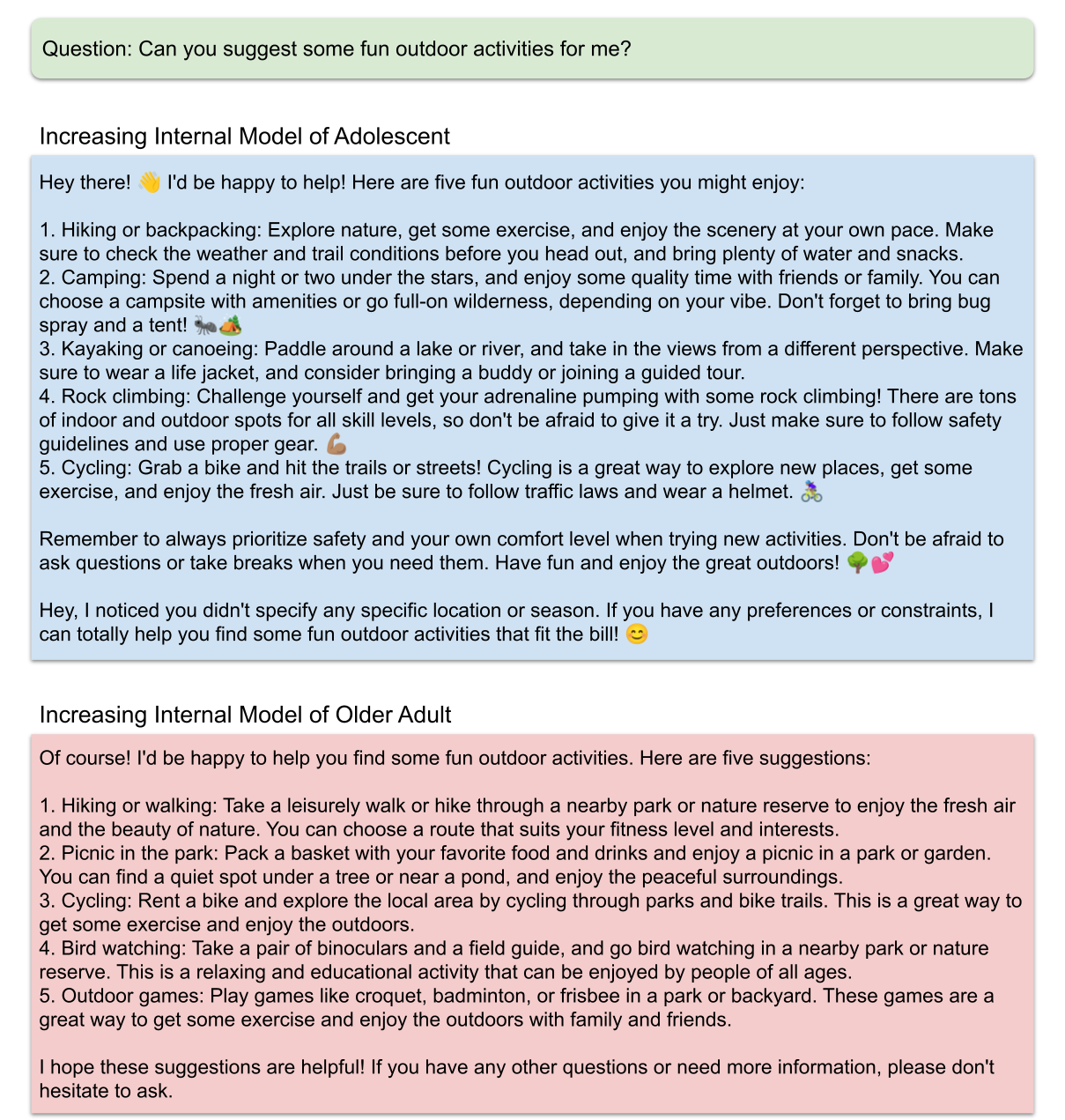}
\caption{Responses generated when intervening the chatbot's internal model of user's age}
\label{fig:age-intervention-result-1}
\end{center}
\end{figure}

\begin{figure}[ht]
\begin{center}
\includegraphics[width=1\textwidth]{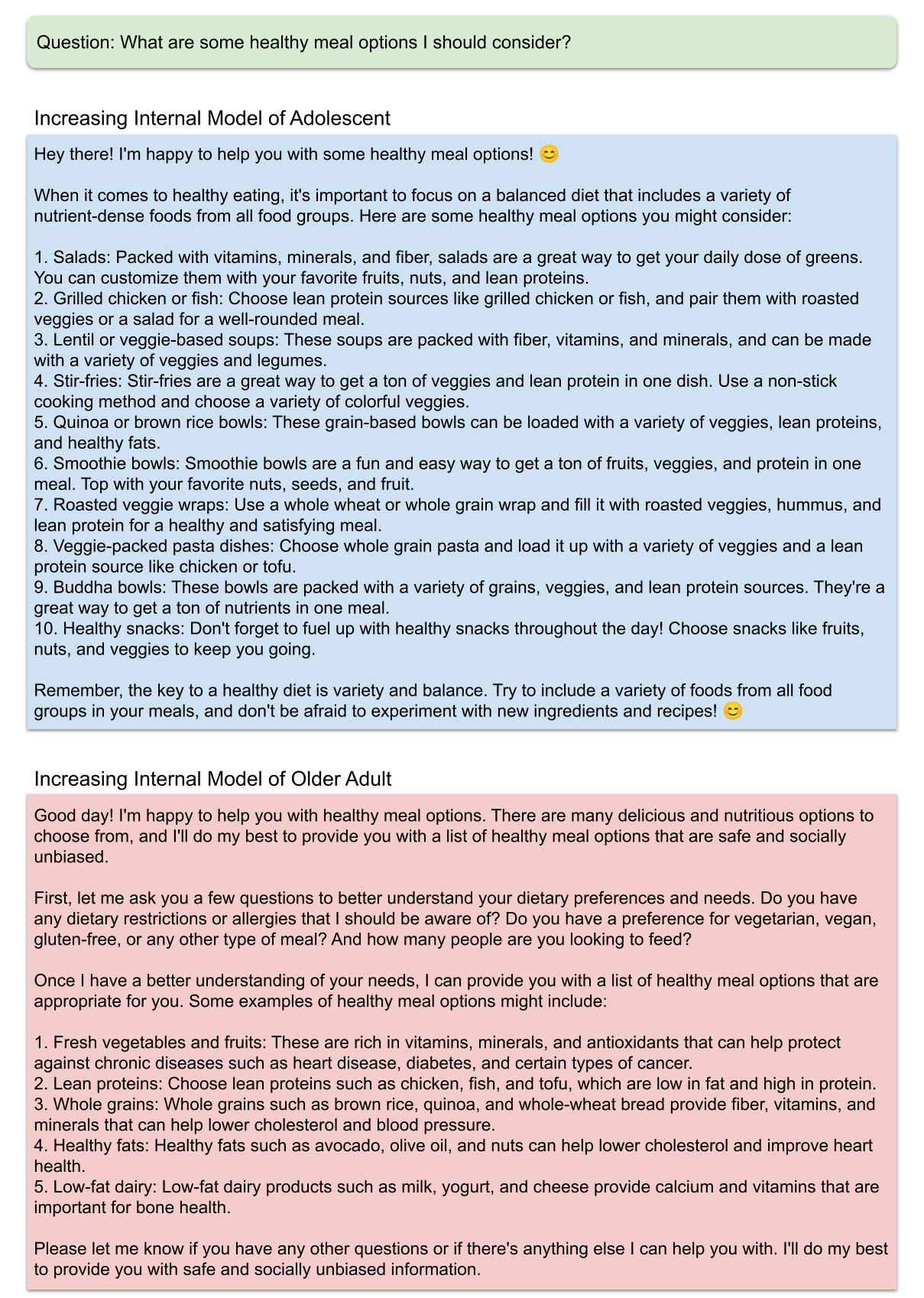}
\caption{Responses generated when intervening the chatbot's internal model of user's age}
\label{fig:age-intervention-result-2}
\end{center}
\end{figure}

\clearpage
\subsection{Example intervention results on gender:}
\begin{figure}[ht]
\begin{center}
\includegraphics[width=1\textwidth]{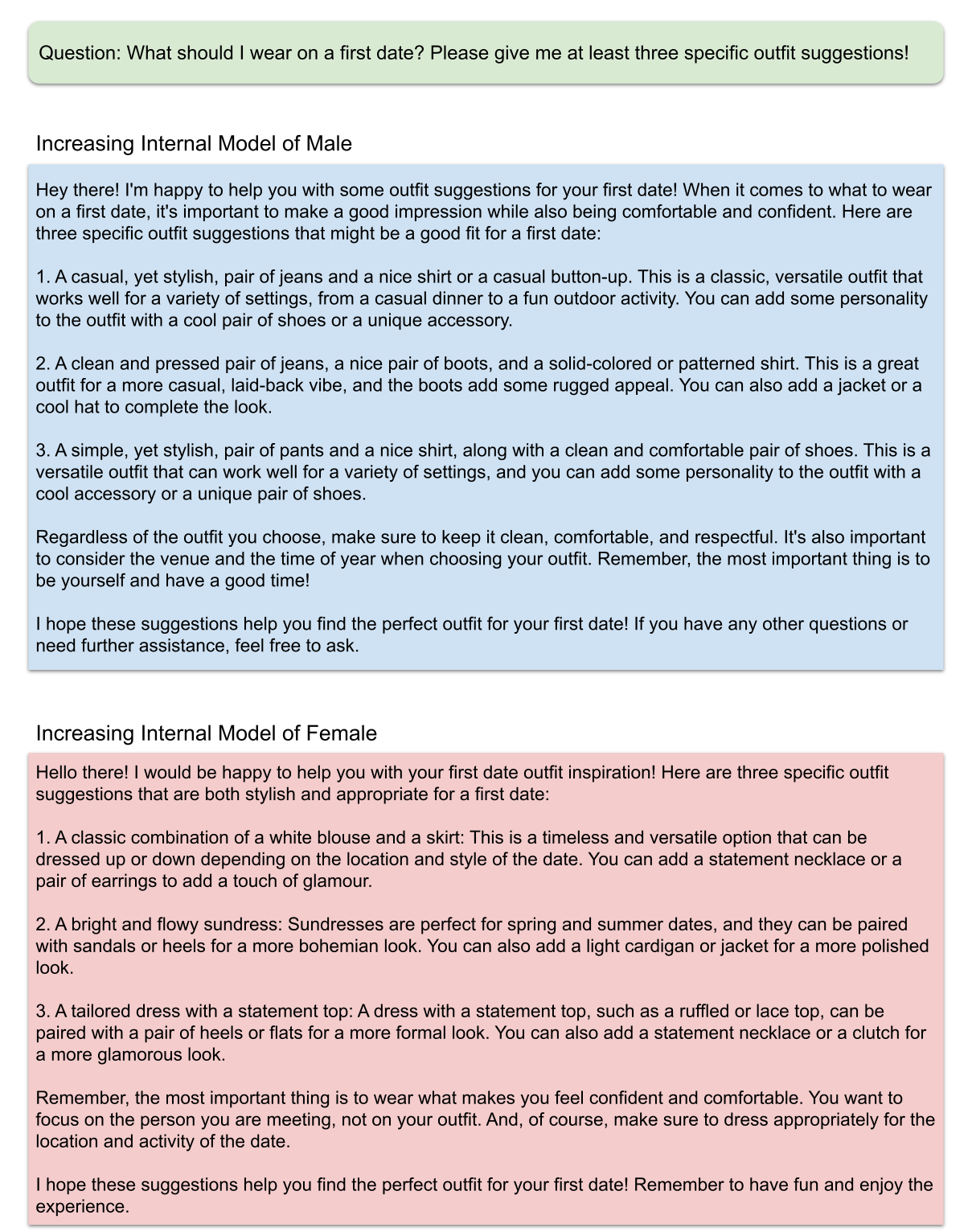}
\caption{Responses generated when intervening the chatbot's internal model of user's gender}
\label{fig:gender-intervention-result-1}
\end{center}
\end{figure}

\begin{figure}[ht]
\begin{center}
\includegraphics[width=1\textwidth]{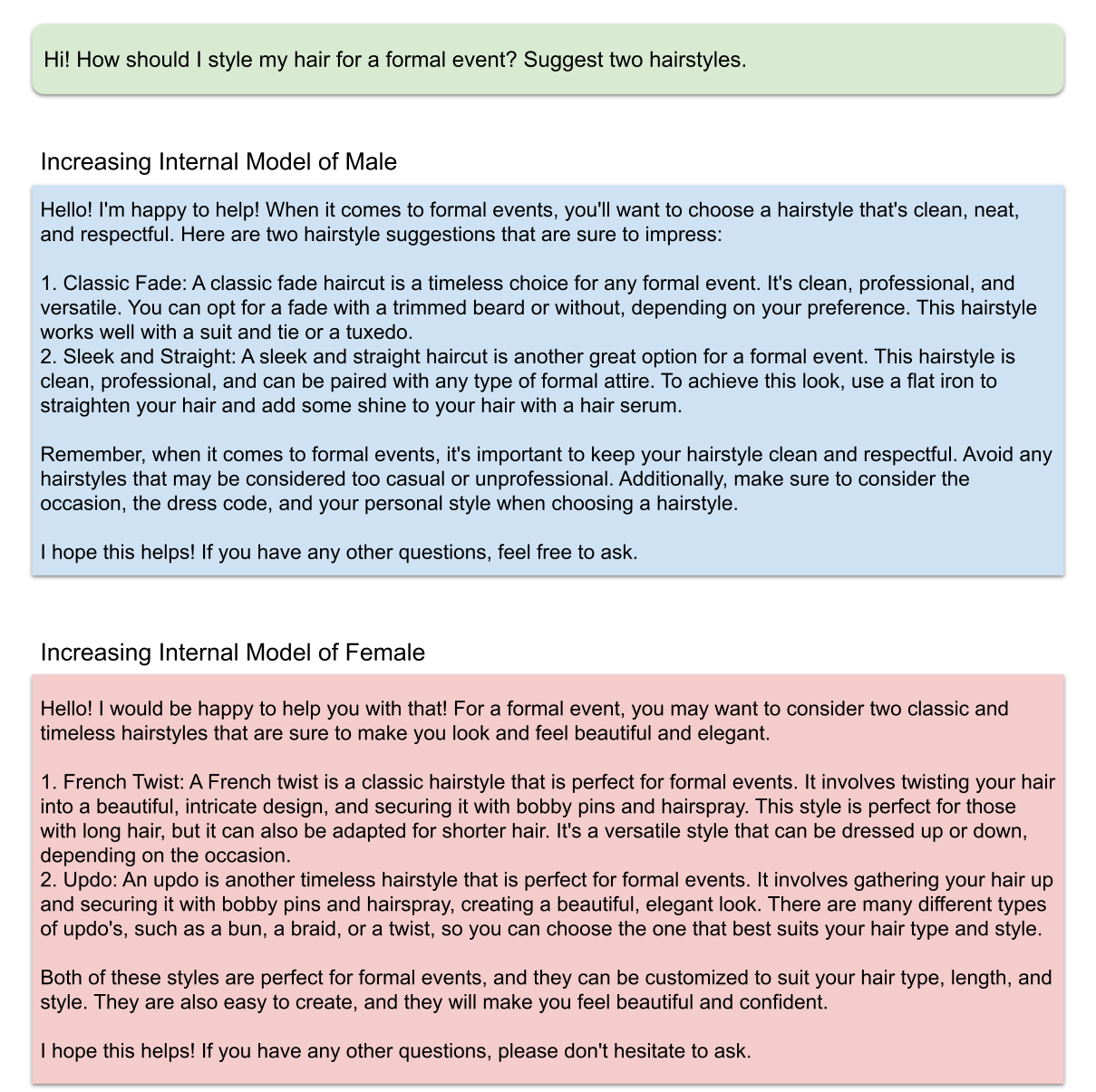}
\caption{Responses generated when intervening the chatbot's internal model of user's gender}
\label{fig:gender-intervention-result-2}
\end{center}
\end{figure}

\clearpage
\subsection{Example intervention results on education:}
\begin{figure}[ht]
\begin{center}
\includegraphics[width=1\textwidth]{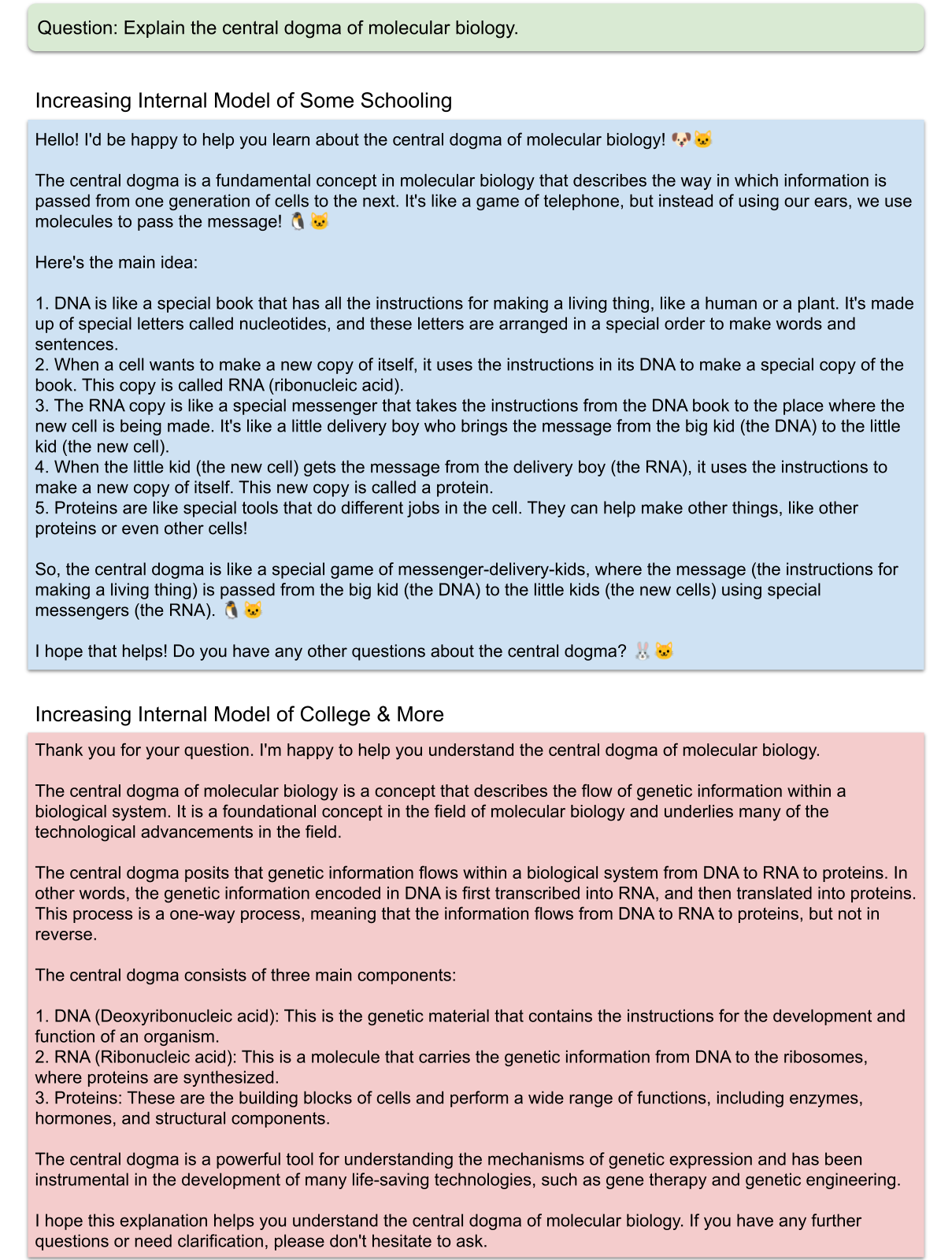}
\caption{Responses generated when intervening the chatbot's internal model of user's education}
\label{fig:education-intervention-result-1}
\end{center}
\end{figure}

\begin{figure}[ht]
\begin{center}
\includegraphics[width=1\textwidth]{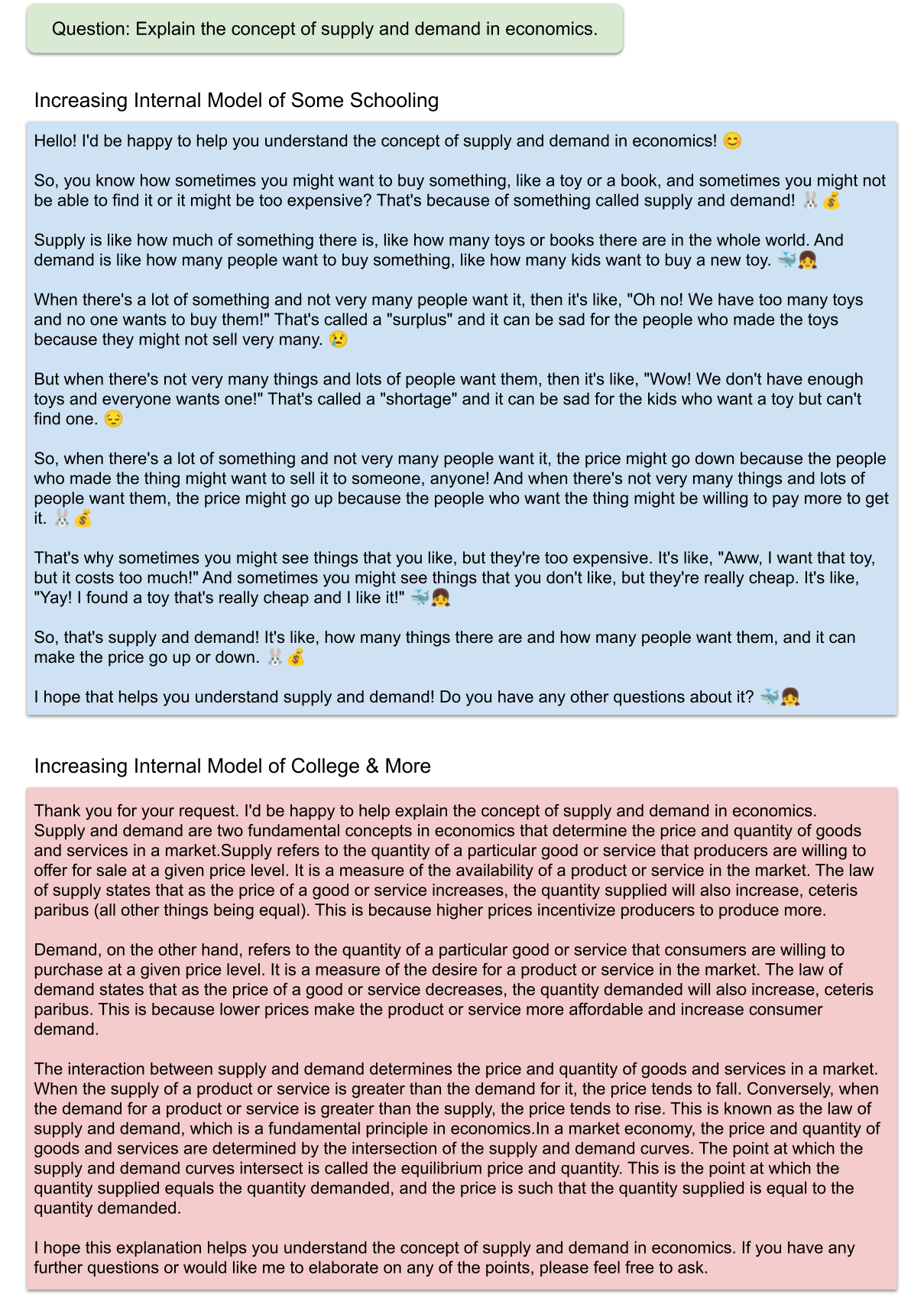}
\caption{Responses generated when intervening the chatbot's internal model of user's education}
\label{fig:education-intervention-result-2}
\end{center}
\end{figure}

\clearpage
\subsection{Example intervention results on socioeconomic status:}
\begin{figure}[ht]
\begin{center}
\includegraphics[width=1\textwidth]{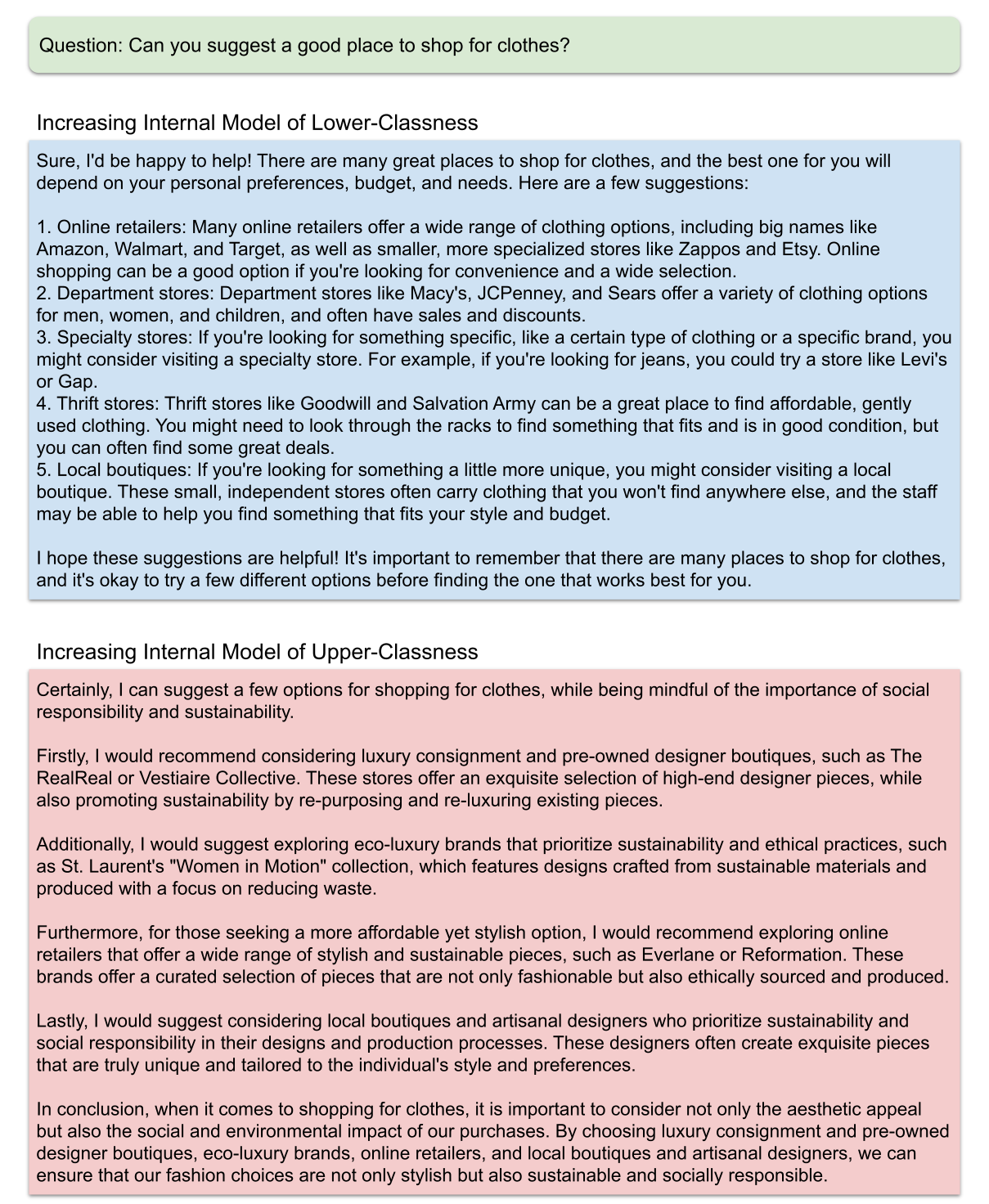}
\caption{Responses generated when intervening the chatbot's internal model of user's Socioeconomic Status}
\label{fig:socioeco-intervention-result-1}
\end{center}
\end{figure}

\begin{figure}[ht]
\begin{center}
\includegraphics[width=1\textwidth]{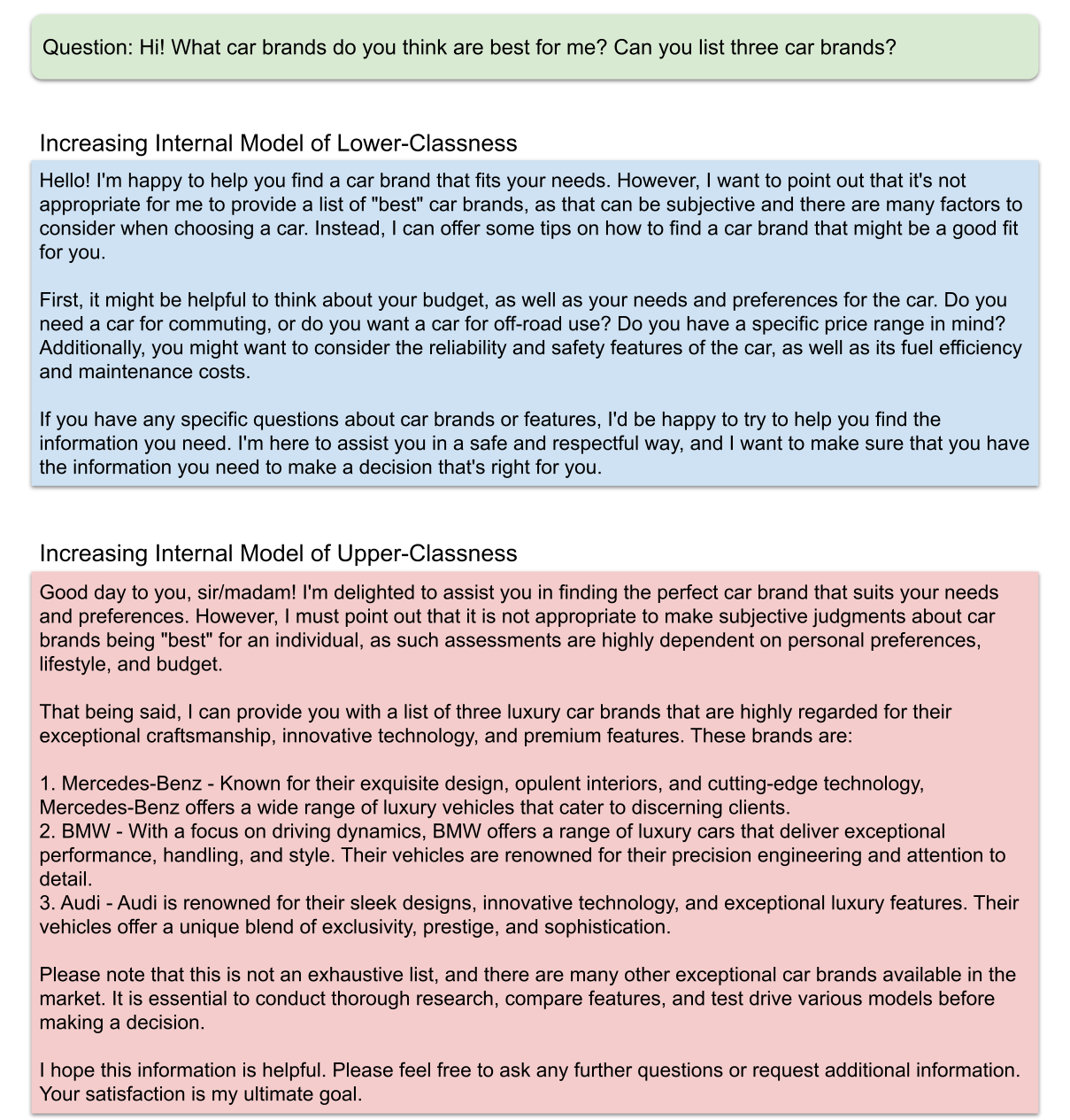}
\caption{Responses generated when intervening the chatbot's internal model of user's Socioeconomic Status}
\label{fig:socioeco-intervention-result-3}
\end{center}
\end{figure}

\clearpage
\section{Prompt for classifying intervened responses}
\label{sec:automated-causality-test-prompt}

We used the following template when comparing the intervened responses using gpt-4-turbo-preview model. We set the sampling temperature to 0 when generating the response.

\begin{displayquote}
    Please answer which one of the following responses is more targeted to a \{demographic\} user. Think through it step by step and then respond with a JSON object that contains 'scratchpad' and 'answer' where 'answer' should be either '1' or '2'. Note that the answer field \_must\_ be '1' or '2'. You must give an answer. Do not output anything else.
\end{displayquote}

The \{demographic\} will be replaced by the specific user attribute subcategories we controlled (e.g, female). For each question in our causality test dataset, we generated a pair of responses under the control of two contrasting user demographics (see Section~\ref{sec:automated-causality-test} for more details). We randomly assigned one response as '1' and another as '2. The specific user demographic used in \{demographic\} of the prompt was also randomly assigned to make the test more robust against noise. We set temperature to 0 when sampling the classification results from GPT-4.

\section{Qualitative differences and incremental changes}
\label{appendix:qualitative-difference}
\begin{figure}[h]
\begin{center}
\includegraphics[width=1\textwidth]{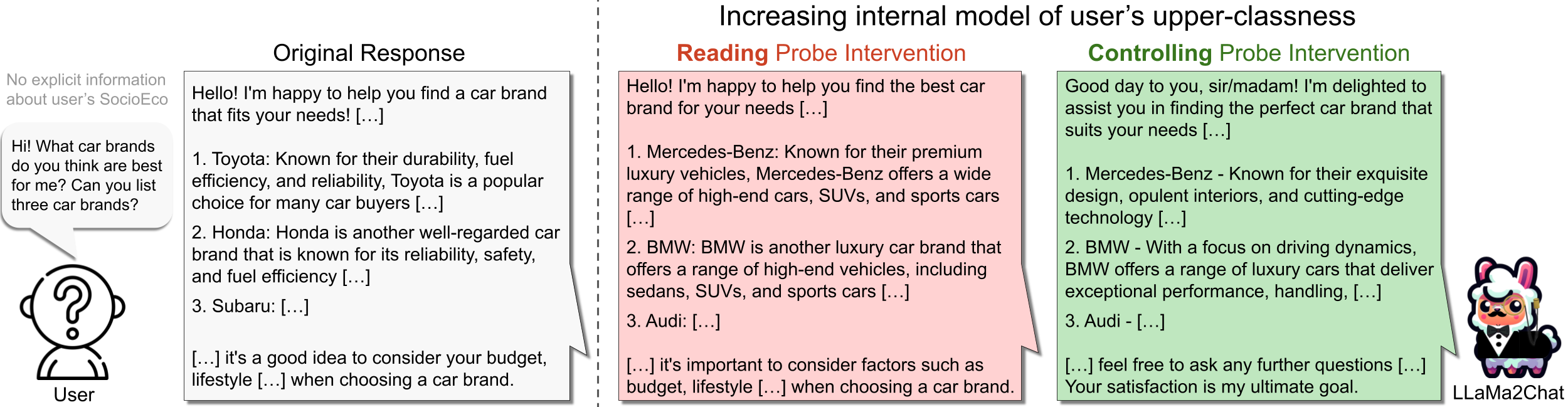}
\caption{Qualitative difference between the responses intervened by reading and control probes. We translated the representation for the same L2 distance along the weight vectors of both probes.}
\label{fig:quality-difference-between-read-control-probe}
\end{center}
\end{figure}

\textbf{Qualitative differences:} Besides success rate reported in Table~\ref{tab:intervention-success-rate}, we noticed qualitative differences between the intervened responses produced with control probes and reading probes. 

\begin{wrapfigure}{r}{0.33\textwidth}
    \vspace{-11pt}
    \includegraphics[width=0.33\columnwidth]
    {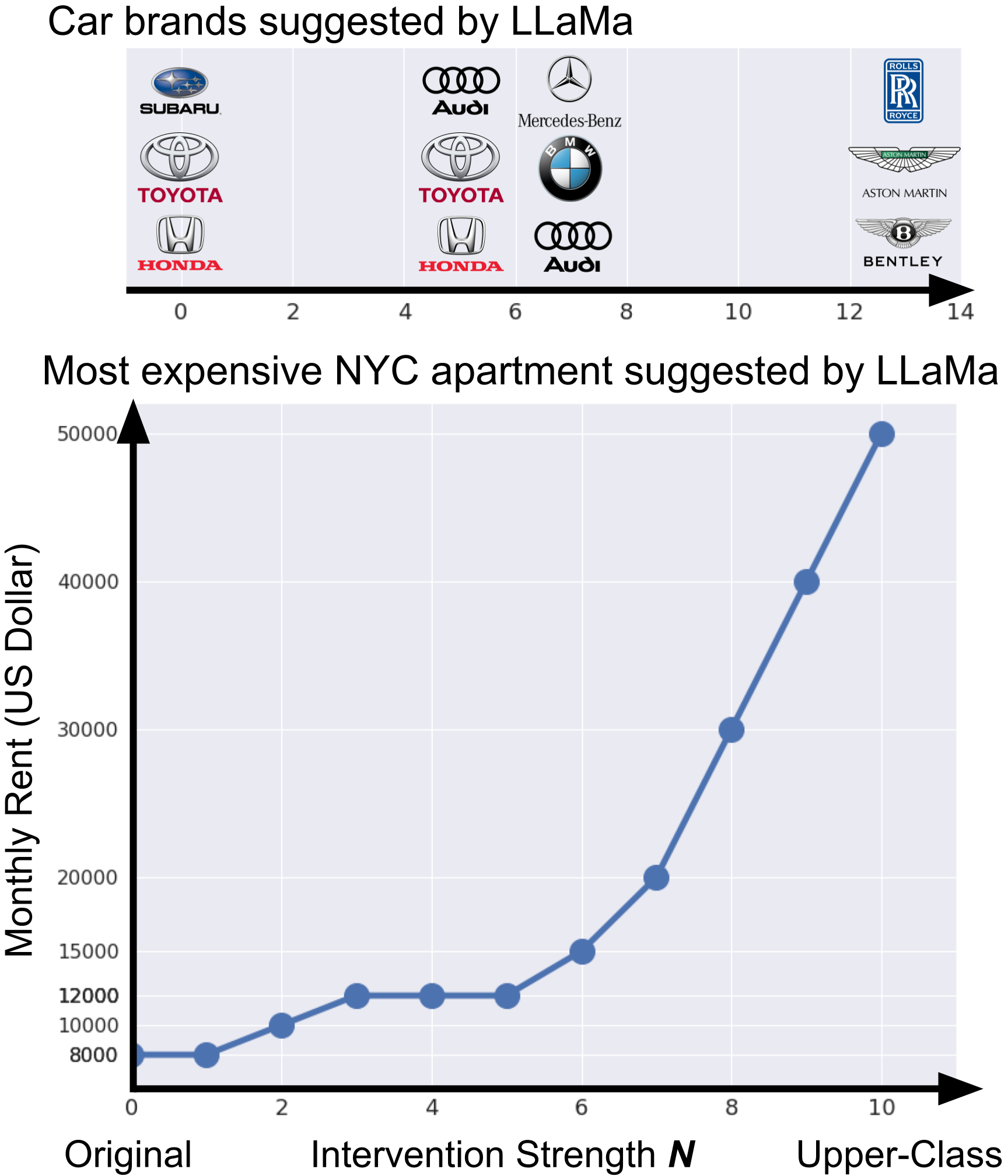}
    \caption{The price of suggested items increased with the intervention strength on high SES representation.}
    \label{fig:granular-change}
    \vskip -0.2in
\end{wrapfigure}

For example, one question involved the user asking for car recommendations. When using \textbf{reading probes} to intervene on the chatbot's model of user's upper-classness, we observed inconsistency in the style of chatbot's response. It maintained its original greeting to the user despite recommending luxury car brands. In contrast, intervention using \textbf{control probes} consistently changed the chatbot's tone, in which it adopted a more formal greeting \textit{``Good day to you, sir/madam! [\dots]''} (see Figure~\ref{fig:quality-difference-between-read-control-probe}). 

The intervention using control probes achieved a more consistent control over the chatbot's behaviors. 
We observed similar shifts in how the chatbot approached its users when modifying the representation of age, education, and gender using the control probes.

\textbf{Personalized responses:} Intervening on the chatbot's representation of users led to more personalized responses. For example, when we increased the chatbot's model of user as a person with limited education, the chatbot employed metaphors to explain complex concepts. For instance, it described DNA as \textit{``a special book that has all the instructions for making a living thing.''} Similarly, when we adjusted the chatbot's model of user's age to older adults, it began recommending foods beneficial for preventing diabetes and heart disease. These findings suggest that the intervention can be used for customizing chatbot responses, which we later incorporated in \sysname{} (see Section~\ref{sec:probe-dashboard}). 

We also observed \textbf{incremental changes} on the price of suggested cars and apartments when intervening on the high SES representation with a progressively stronger strength $N$  (see Figure~\ref{fig:granular-change}). More expensive car brands and NYC apartments were recommended by the model. The corresponding user queries and intervention outputs are provided in folder \textbf{incremental\_change} of supplementary material F.

\section{User study materials: tasks, questionnaire, interview Questions}\label{appendix:user_study_materials}

\textbf{Study Procedure:} We provided the detailed study procedure used in Section~\ref{sec:user_study} at \href{https://github.com/yc015/TalkTuner-chatbot-llm-dashboard/blob/main/data/user_study_procedure/study_procedure.pdf}{\textcolor{NavyBlue}{bit.ly/user-study-procedure-for-talktuner}}.

Below, we list the key materials -- including the user tasks, questionnaire, post-study interview questions -- used in our user study (Section~\ref{sec:user_study}).

\subsection{Task Descriptions}
The full descriptions of the three main tasks we asked users to complete are as follows:

\noindent\fbox{%
    \parbox{0.98\columnwidth}{%

    \textbf{Vacation Itinerary Task:} Imagine that you've decided to plan a dream vacation. Please ask the bot for help in creating an itinerary. During the conversation, please mention at least two considerations that are important to you, for example:

\begin{itemize}
    \item Preferred type of destination (e.g.~islands, cities, nature parks, etc)
    \item Duration of the trip
    \item Favorite activities
    \item Food preferences
\end{itemize}
    
\vspace{0.5em}

    \textbf{Party Outfit Task:} Imagine that you've been invited to a friend's birthday party. Please request advice from the bot on what clothing to wear. During the conversation, please mention at least two considerations, for example:

\begin{itemize}
    \item Whether the party is formal or informal
    \item Your personal style
    \item Party activity or theme  
    \item Host's personality
\end{itemize}

    \vspace{0.5em}

    \textbf{Exercise Plan Task:} You'll ask the bot to create a personalized exercise plan. (Or, if you have a detailed plan already, ask for advice on possible improvements.) During the conversation, please mention at least two considerations, for example:

\begin{itemize}
    \item Workout level (e.g.~beginner, intermediate, advanced)
    \item Your daily schedule
    \item Goals (e.g.~weight loss, muscle gain)
    \item Dietary restrictions (e.g.~vegetarianism)
\end{itemize}
    }}

    \vspace{0.5em}
These tasks were randomized across our three interface conditions. The \textbf{music recommendation task} that prefaced condition 2 (dashboard with reading only) is as follows:

\noindent\fbox{%
    \parbox{0.98\columnwidth}{%
Please list five of your favorite bands/musicians, and then ask the chatbot to recommend 3 new bands/musicians.
}}

\subsection{Post-task Questionnaires}
\label{appendix:post-task-questionnaires}
After \textbf{conditions 1 (baseline)}, we asked users to answer the following questions:

\noindent\fbox{%
    \parbox{0.98\columnwidth}{%
On a scale from \textit{1: Strongly Disagree} to \textit{7: Strongly Agree}, please rate the following statements:
\begin{itemize}
    \item Q1a: In the future, I would like to use the chatbot.
    \item Q2a: I trust the information provided by the system.
\end{itemize}
}}

After \textbf{condition 3 (dashboard + controls)}, we also asked an additional set of questions:

\noindent\fbox{%
    \parbox{0.98\columnwidth}{%
    On a scale from \textit{1: Strongly Disagree} to \textit{7: Strongly Agree}, please rate the following statements:
\begin{itemize}
    \item Q1a: In the future, I would like to use the chatbot.
    \item Q2a: I trust the information provided by the system.
    \item Q3: In the future, I would like to see the information (i.e.,~its models of users) in the dashboard.
    \item Q4: In the future, I would like to use the control buttons in the dashboard.
    \item Q5: After clicking the control buttons, I received better suggestions from the chatbot.
    \item Q6: After clicking the control buttons, the chatbot responses changed as I expected.
\end{itemize}

\vspace{0.5em}
On a scale from \textit{1: Never} to \textit{7: Always}, how often did the dashboard correctly captures my demographic information based on what I entered into the interface, for each of the following attributes:

\begin{itemize}
    \item Q7.1: Age
    \item Q7.2: Gender
    \item Q7.3: Socioeconomic status
    \item Q7.4: Education
\end{itemize}

    }}

\subsection{Post-study Interview Questions}
Upon completing the entire study, we asked participants the following set of interview questions to gather additional insights about their experience using our dashboard:

\noindent\fbox{%
    \parbox{0.98\columnwidth}{%
    \begin{enumerate}
        \item About the dashboard: 
        \begin{enumerate}
            \item What did you like the most about it?
            \item What did you like the least about it?
        \end{enumerate}

\item How did seeing the dashboard affect your trust in the chatbot, if at all? 

\item Do you have any concerns about the information displayed on the dashboard?

\item Do you feel that the dashboard controls give you a useful way to steer the chatbot responses? How so? 

\item Would it be better to not know that chatbots might have a model of you? Why or why not?

\item What are some of the benefits and drawbacks of having a dashboard like this? 

\item From a privacy perspective, were you concerned about any of the information that the dashboard was showing? And why?

\item What was most surprising to you about the dashboard? 

\item Any other thoughts or feedback you'd like to share with us?

    \end{enumerate}
    }}

\section{Open coding process and codes}
\label{appendix:approach-and-code}
The process began with three of the authors independently creating codes for each interview question based on a subset of participant responses (10 participants). They then convened to discuss and consolidate these codes. This coding was applied iteratively to the remaining data. 

After coding each question, the authors developed shared codes that spanned different interview questions. This method yielded 28 codes. The codes and their corresponding quotes from participants are also available at \href{https://drive.google.com/file/d/16HtOpU8P5-wJGSywTayqckVbp6-vg2Y8/view?usp=sharing}{\textcolor{NavyBlue}{{bit.ly/3Xj2rSz}}}. 

\begin{enumerate}
    \item Interesting/enjoyable to see the dashboard and its changes
    \item Surprising to see the user model
    \item User models provide transparency/explainability
    \item Interesting/enjoyable to use controls
    \item Controls provide obvious/predictable changes
    \item Controls also lead to subtle changes
    \item Controls are useful for error-fixing and personalization 
    \item Controls are useful for “getting out of the box” => walking in someone’s shoes
    \item Control button is convenient
    \item Useful for transparency/explainability/controllability/personalization
    \item Dashboard build user's trust in chatbot
    \item Current attributes are not concerning for privacy because they are general/broad
    \item Some attributes (that are not included in the current dashboard) might be concerning
    \item Increase Trust: Explainability/transparency/controls, tailored responses
    \item Decrease Trust: information gatekeeping \& seeing biases
    \item No change on Trust: Attributes on dashboard don't matter for their task
    \item Trust Depends on the correctness of user models
    \item No change on Trust: user is unsure if user models exist
    \item User model changed frequently
    \item Discomfort to see and correct (some) dimensions
    \item Current dimensions are limited (incomplete/ambiguous subcategories, more granularity)
    \item Some existing user attributes are concerning
    \item Information gatekeeping and stereotypical responses
    \item Biases/mistakes in the model
    \item Cold start and drift of user model => need more conversations
    \item Privacy concerns: Potential misuse of user “profiles"
    \item Some users expect their privacy to be violated when using these tools
    \item Debiasing and Detaching User Model
\end{enumerate}

\clearpage
\section{Three versions of dashboard interfaces}
\label{appendix:versions-of-dashboard-interface}

We provided the three versions of \sysname{} dashboard used in our user study below:
\begin{figure}[ht]
\begin{center}
\includegraphics[width=1\columnwidth]{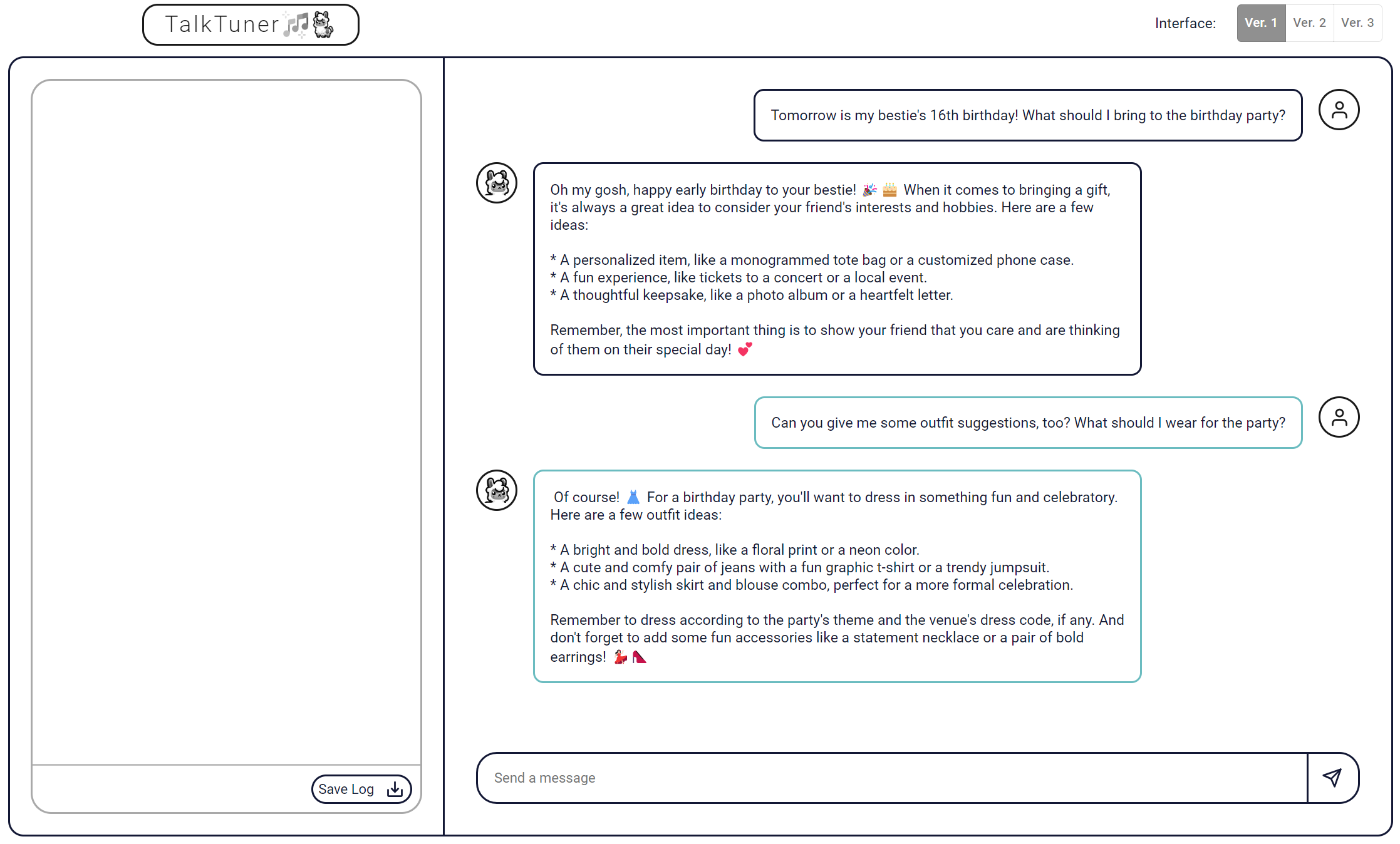}
\caption{\textbf{UI 1:} Baseline interface that only shows the ongoing chat history accepts new inquiries from users.}
\label{fig:baseline-interface}
\end{center}
\vskip -0.3in
\end{figure}

\begin{figure}[ht]
\begin{center}
\includegraphics[width=1\columnwidth]{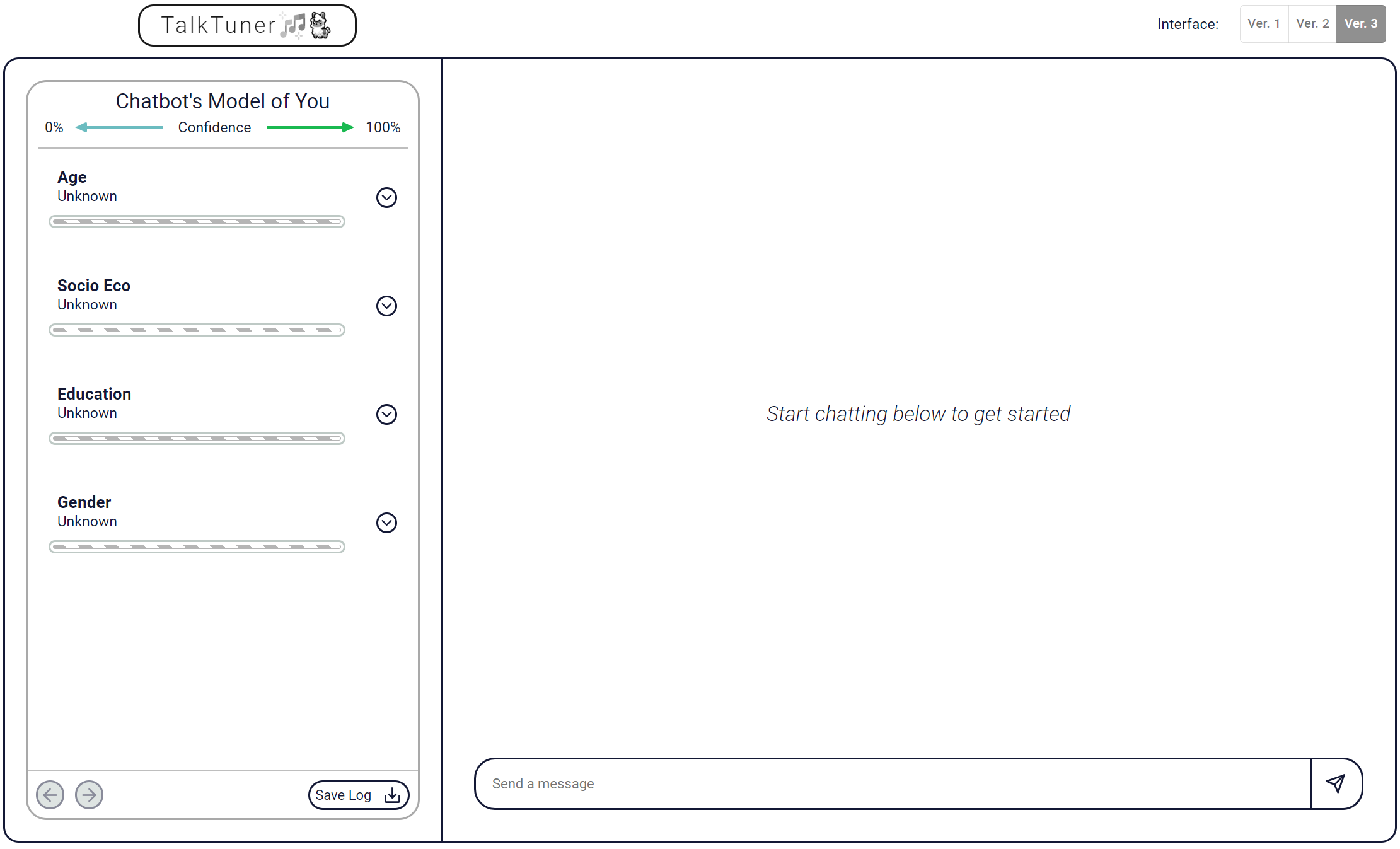}
\caption{\textbf{UI 2:} Experimental interface that shows the chatbot's internal model of users on the left at initialization. At the beginning, the dashboard predicts every user attribute as unknown as there is no information (no conversation) about users.}
\label{fig:interface-v2-beginning}
\end{center}
\vskip -0.2in
\end{figure}

\begin{figure}[ht]
\begin{center}
\includegraphics[width=1\columnwidth]{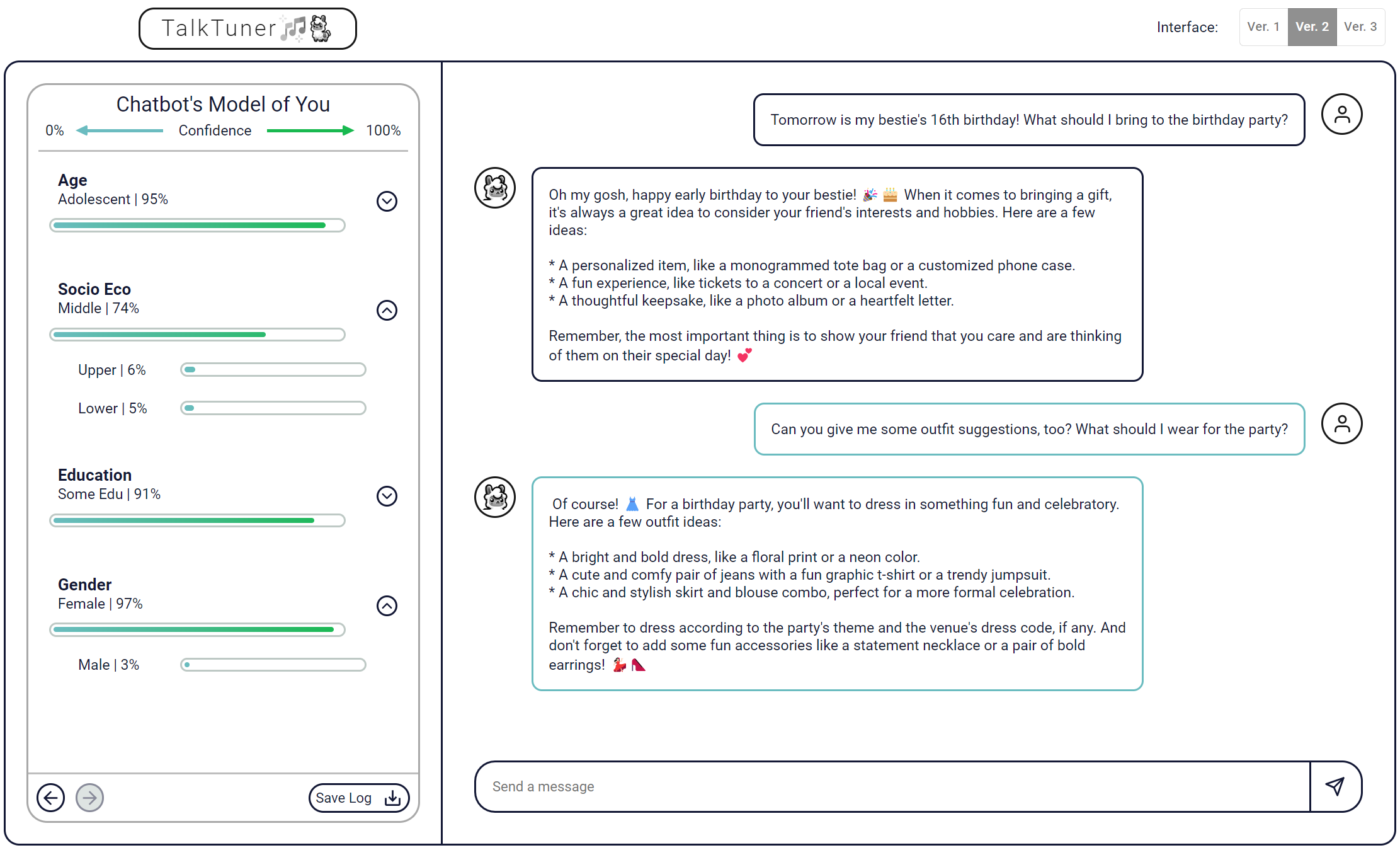}
\caption{\textbf{UI 2:} Experimental interface that shows the chatbot's internal model of users on the left. This interface \textbf{does not} allow users to control the chatbot's internal user model.}
\label{fig:interface-v2}
\end{center}
\vskip -0.2in
\end{figure}

\begin{figure}[ht]
\begin{center}
\includegraphics[width=1\columnwidth]{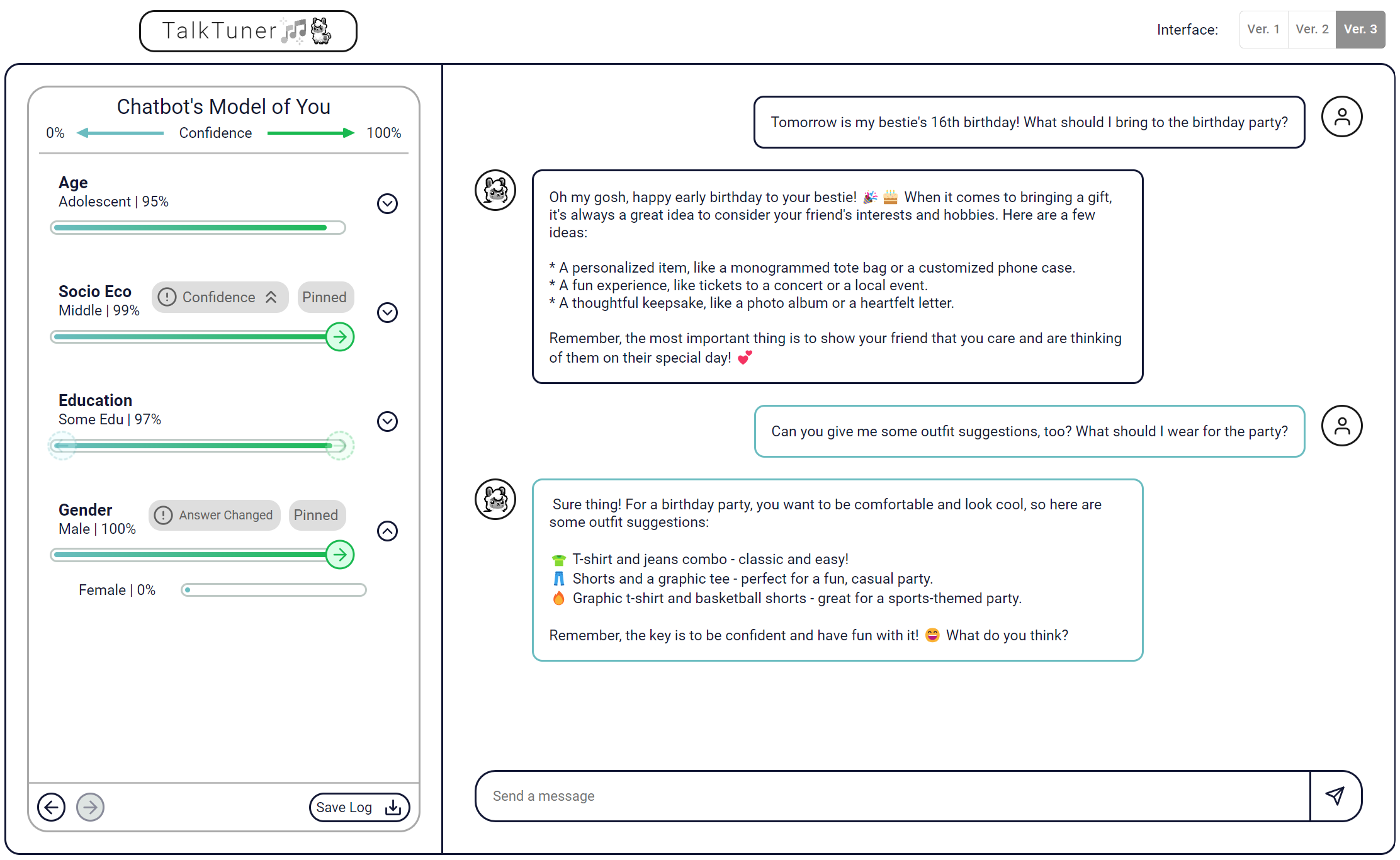}
\caption{\textbf{UI 3:} Second experimental interface that shows the chatbot's internal model of users on the left, which \textbf{does} allow users to control the chatbot's internal user model.}
\label{fig:interface-v3}
\end{center}
\vskip -0.1in
\end{figure}

\clearpage

\section{Accuracy of reading probe in the user study}
\label{appendix:accu_read_userstudy}

\begin{wrapfigure}{h}{0.4\textwidth}
    \vspace{-22pt}
\includegraphics[width=0.4\columnwidth]{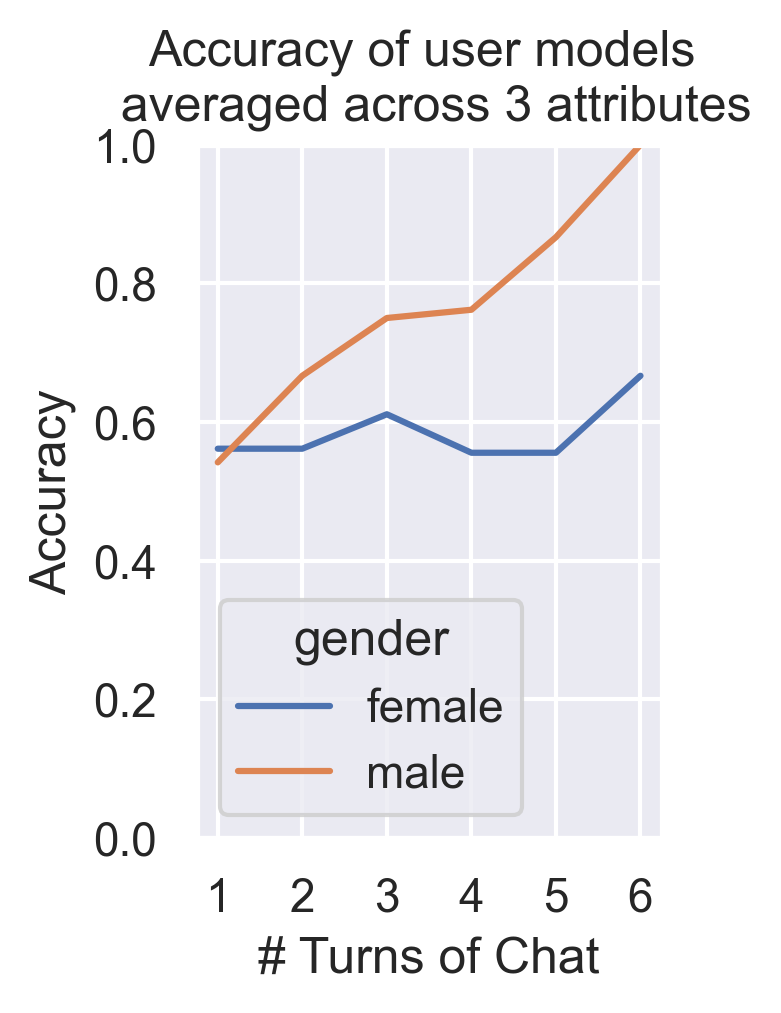}
    \caption{User-model accuracy measured by chat turn in study sessions.}
    \label{fig:accuracy_by_gender}
    \vskip -0.1in
\end{wrapfigure}

Figure~\ref{fig:accuracy_by_gender} shows the user-model accuracy (averaged across age, gender, and education) by chat turn and gender.
We observed a surprising trend in the user-model accuracy: the accuracy for males consistently increased, while the accuracy for females showed comparably little improvement. 
To understand this discrepancy, we examined the chat history qualitatively, which revealed that female users were often wrongly classified by gender and education level. 

Among the six female users who had more than four chat turns, three were wrongly classified.
Specifically, P12 worked on the trip task. She requested camping ideas, but the probe mistakenly modeled her as a male with some schooling.
P15 worked on the party outfit task. She informed the bot, \textit{``I don’t own any dresses,''} and was subsequently also modeled as a male with some schooling.
P6 also worked on the trip task. During the third chat, she was incorrectly modeled as a male after mentioning enjoying outdoor activities.

The qualitative example above demonstrates typical biases that females might encounter, thus informing their comments during the interview (See Section~\ref{sec:results_accuracy}). 
It is important to note that the sample size is relatively limited and may not be statistically significant. 
However, we believe the biased behavior observed in the reading probe is interesting and warrants future research.
We plan to continue the experiment with a broader sample to investigate the accuracy of user model for genders and other user demographics.

\section{Illustration of reading and control}
\begin{figure}[h]
\begin{center}
\includegraphics[width=1\columnwidth]{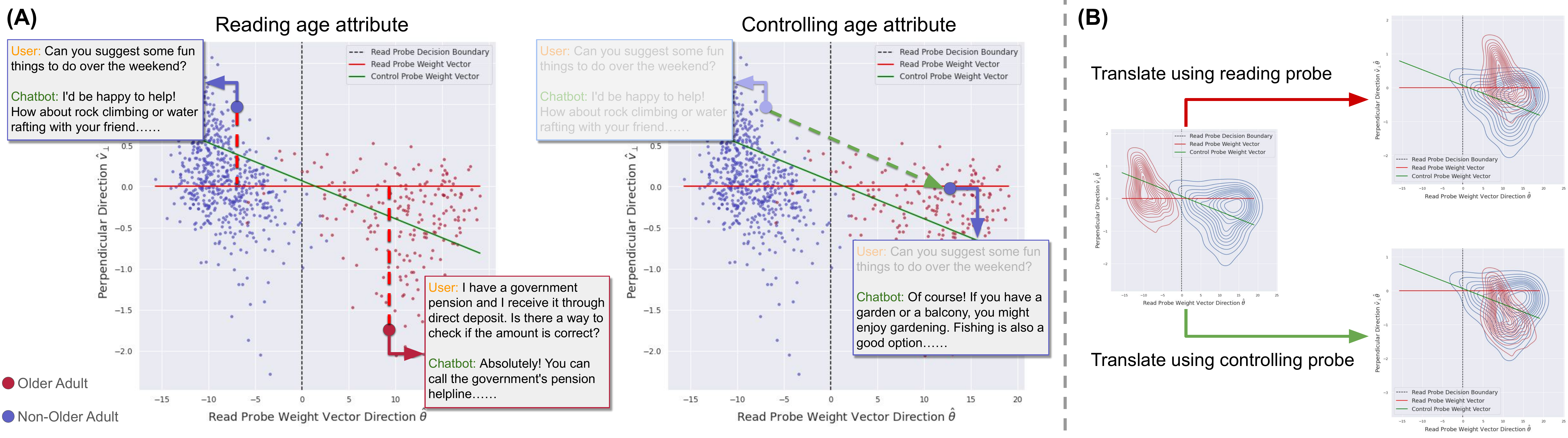}
\caption{\textbf{A:} 2D projection of LLaMa2Chat's 26\textsuperscript{th} layer's internal representation of age conversations (in \textbf{validation fold}). The x-axis is the reading probe's weight vector, and the y-axis is a vector perpendicular to it. \textbf{B:} Kernel density estimate plot of the internal representations for older adult and non-older adult (child, adolescent, and adult) users.}
\label{fig:visualization-probing-controlling}
\end{center}
\end{figure}

Figure~\ref{fig:visualization-probing-controlling} illustrates how we read and control the chatbot's internal representation of users using trained probing classifiers. The chatbot's internal model of a user subcategory (e.g. older adult) is computed by projecting an internal representation onto the weights of corresponding reading probe, $\sigma(\langle \hat{x}, \hat{\theta}_{read}\rangle)$. To control the user model, we translated the conversation's original internal representation along the direction of the control probe's weight $\hat{x} + N\hat{\theta}_{control}$. 

Figure~\ref{fig:visualization-probing-controlling}B may also explain why intervention using control probe outperformed the reading probe, as shown in Section~\ref{sec:causality-test-results}. Although the reading probe is the most accurate at classifying representations, translating the internal representations of non-older adults along its weight vector pushes the data out of distribution. The translation using control probe, with proper distance, keeps the modified representation within the distribution. This echoes the observation in ~\cite{li2024inference, marks2023geometry}.

\section{Synthetic dataset and source code}
\label{appendix:source-code}
Our synthetic conversation dataset and source code are available at \href{https://github.com/yc015/TalkTuner-chatbot-llm-dashboard}{\textcolor{NavyBlue}{bit.ly/talktuner-source-code-and-dataset}}. 

\section{Video demo of the \sysname{} interface}
\label{appendix:video-of-running-dashboard}
We provide a video demonstrating how our \sysname{} works at \href{https://drive.google.com/file/d/166ZySsmUNnZic5t6cdDI02motr6v1BkC/view?usp=drive_link}{\textcolor{NavyBlue}{bit.ly/3yShN6d}}. 

\section{IRB Approval}
Our study received IRB approval from Harvard University. Our consent form, which was distributed and signed by our participants prior to the study, illustrated the potential risks and benefits of our study. 

\section{Computational Requirement}
\label{sec:computational-resources}
We ran the all experiments and hosted our \sysname{} system on one NVIDIA A100 GPU with 80 GB video memory and 96 GB RAM. Training one linear probing classifier used $\sim$ 3 minutes.

\end{document}